\documentclass{article}

\usepackage[preprint,nonatbib]{neurips_2019}

\usepackage[utf8]{inputenc} 
\usepackage[T1]{fontenc}    
\usepackage{hyperref}       
\usepackage{url}            
\usepackage{booktabs}       
\usepackage{amsfonts}       
\usepackage{nicefrac}       
\usepackage{microtype}      

\usepackage{amsmath}
\usepackage{graphicx}
\usepackage{algorithm}
\usepackage{algorithmic}
\usepackage{bbm}
\usepackage{xcolor}
\usepackage{pifont}
\usepackage{fancyvrb}

\usepackage{booktabs}		
\usepackage{multirow}

\DeclareMathOperator{\st}{s.t.}
\DeclareMathOperator{\argmin}{argmin}

\newcommand{\ignore}[1]{}

\newcommand{\cmark}{\ding{51}}
\newcommand{\xmark}{\ding{55}}
\newcommand{\nCheck}{\textcolor{red!90!black}{\xmark}}
\newcommand{\pCheck}{\textcolor{green!70!black}{\cmark}}
\newcommand{\norm}[1]{\left\Vert#1\right\Vert}
\newcommand{\setR}{\mathbb{R}}

\title{GENO -- GENeric Optimization\\ for Classical Machine Learning}

\author{
  S\"oren Laue\\
  Friedrich-Schiller-Universit\"at Jena\\
  Germany\\
  \texttt{soeren.laue@uni-jena.de} \\
  \And Matthias Mitterreiter\\
  Friedrich-Schiller-Universit\"at Jena\\
  Germany\\
  \texttt{matthias.mitterreiter@uni-jena.de} \\
  \And Joachim Giesen\\
  Friedrich-Schiller-Universit\"at Jena\\
  Germany\\
  \texttt{joachim.giesen@uni-jena.de} \\
}

\begin{document}

\maketitle

\begin{abstract}
  Although optimization is the longstanding algorithmic backbone of
  machine learning, new models still require the time-consuming
  implementation of new solvers. As a result, there are thousands of
  implementations of optimization algorithms for machine learning
  problems. A natural question is, if it is always necessary to
  implement a new solver, or if there is one algorithm that is
  sufficient for most models. Common belief suggests that such a
  one-algorithm-fits-all approach cannot work, because this algorithm
  cannot exploit model specific structure and thus cannot be efficient
  and robust on a wide variety of problems. Here, we challenge this
  common belief. We have designed and implemented the optimization
  framework GENO (GENeric Optimization) that combines a modeling
  language with a generic solver. GENO generates a solver from the
  declarative specification of an optimization problem class. The
  framework is flexible enough to encompass most of the classical
  machine learning problems. We show on a wide variety of classical
  but also some recently suggested problems that the automatically
  generated solvers are (1) as efficient as well-engineered
  specialized solvers, (2) more efficient by a decent margin than
  recent state-of-the-art solvers, and (3) orders of magnitude more
  efficient than classical modeling language plus solver approaches.
\end{abstract}

\section{Introduction}

Optimization is at the core of machine learning and many other fields
of applied research, for instance operations research, optimal
control, and deep learning. The latter fields have embraced frameworks
that combine a modeling language with only a few optimization solvers;
interior point solvers in operations research and stochastic gradient
descent (SGD) and variants thereof in deep learning frameworks like
TensorFlow, PyTorch, or Caffe. That is in stark contrast to classical
(i.e., non-deep) machine learning, where new problems are often
accompanied by new optimization algorithms and their
implementation. However, designing and implementing optimization
algorithms is still a time-consuming and error-prone task.

The lack of an optimization framework for classical machine learning
problems can be explained partially by the common belief, that any
efficient solver needs to exploit problem specific structure. Here, we
challenge this common belief.

We introduce GENO (GENeric Optimization), an optimization framework
that allows to state optimization problems in an easy-to-read modeling
language. From the specification an optimizer is automatically
generated by using automatic differentiation on a symbolic level. The
optimizer combines a quasi-Newton solver with an augmented Lagrangian
approach for handling constraints.

Any generic modeling language plus solver approach frees the user from
tedious implementation aspects and allows to focus on modeling aspects
of the problem at hand. However, it is required that the solver is
efficient and accurate. Contrary to common belief, we show here that
the solvers generated by GENO are (1) as efficient as well-engineered,
specialized solvers at the same or better accuracy, (2) more efficient
by a decent margin than recent state-of-the-art solvers, and (3)
orders of magnitude more efficient than classical modeling language
plus solver approaches.

\paragraph{Related work.}

Classical machine learning is typically served by toolboxes like
scikit-learn~\cite{scikit-learn}, Weka~\cite{weka2}, and
MLlib~\cite{mllib}. These toolboxes mainly serve as wrappers for a
collection of well-engineered implementations of standard solvers like
LIBSVM~\cite{ChangL01} for support vector machines or
glmnet~\cite{FriedmanHT09} for generalized linear models. A
disadvantage of the toolbox approach is a lacking of flexibility. An
only slightly changed model, for instance by adding a non-negativity
constraint, might already be missing in the framework.

Modeling languages provide more flexibility since they allow to
specify problems from large problem classes. Popular modeling
languages for optimization are CVX~\cite{cvx,cvx2} for MATLAB and its
Python extension CVXPY~\cite{cvxpy2,cvxpy}, and
JuMP~\cite{jump} which is bound to Julia. In the operations research
community AMPL~\cite{fourerGK03} and GAMS~\cite{Gams} have been used
for many years. All these languages take an instance of an
optimization problem and transform it into some standard form of a
linear program (LP), quadratic program (QP), second-order cone program
(SOCP), or semi-definite program (SDP). The transformed problems is
then addressed by solvers for the corresponding standard
form. However, the transformation into standard form can be
inefficient, because the formal representation in standard form can
grow substantially with the problem size. This representational
inefficiency directly translates into computational inefficiency.

The modeling language plus solver paradigm has been made deployable in
the CVXGEN~\cite{Cvxgen}, QPgen~\cite{qpgen}, and OSQP~\cite{osqpgen}
projects. In these projects code is generated for the specified
problem class. However, the problem dimension and sometimes the
underlying sparsity pattern of the data needs to be fixed. Thus,
the size of the generated code still grows with a growing problem
dimension. All these projects are targeted at embedded systems and are
optimized for small or sparse problems. The underlying solvers are
based on Newton-type methods that solve a Newton system of equations
by direct methods. Solving these systems is efficient only for small
problems or problems where the sparsity structure of the Hessian can
be exploited in the Cholesky factorization.  Neither condition is
typically met in standard machine learning problems.

Deep learning frameworks like TensorFlow~\cite{tf},
PyTorch~\cite{pytorch}, or Caffe~\cite{caffe} are efficient and fairly
flexible. However, they target only deep learning problems that are
typically unconstrained problems that ask to optimize a separable sum
of loss functions. Algorithmically, deep learning frameworks usually
employ some form of stochastic gradient descent
(SGD)~\cite{robbins1951}, the rationale being that computing the full
gradient is too slow and actually not necessary. A drawback of
SGD-type algorithms is that they need careful parameter tuning of, for
instance, the learning rate or, for accelerated SGD, the
momentum. Parameter tuning is a time-consuming and often
data-dependent task. A non-careful choice of these parameters can turn
the algorithm slow or even cause it to diverge. Also, SGD type
algorithms cannot handle constraints.

GENO, the framework that we present here, differs from the standard
modeling language plus solver approach by a much tighter coupling of
the language and the solver. GENO does not transform problem instances
but whole problem classes, including constrained problems, into a very
general standard form. Since the standard form is independent of any
specific problem instance it does not grow for larger instances. GENO
does not require the user to tune parameters and the generated code
is highly efficient.

\begin{table}[h!]
\centering
\caption{Comparison of approaches/frameworks for optimization in
  machine learning.}
\label{tab:advantages}
  \begin{tabular}{p{44mm}ccccc}
    \toprule
      & handwritten & TensorFlow, & Weka, & \multirow{2}{*}{CVXPY} & \multirow{2}{*}{GENO}\\
      & solver & PyTorch  & Scikit-learn\\
    \midrule
      flexible
	& \nCheck & \pCheck & \nCheck & \pCheck & \pCheck \\
    efficient
	& \pCheck & \pCheck & \pCheck & \nCheck & \pCheck \\
    deployable / stand-alone
	& \pCheck & \nCheck & \nCheck & \nCheck & \pCheck \\
      can accommodate constraints
	& \pCheck & \nCheck & \pCheck & \pCheck & \pCheck \\
      parameter free (learning rate, ...) \hspace{-3mm}
	& \nCheck/\pCheck  & \nCheck & \pCheck & \pCheck & \pCheck \\
      allows non-convex problems
	& \pCheck & \pCheck & \pCheck & \nCheck & \pCheck \\
    \bottomrule
  \end{tabular}
\end{table}

\section{The GENO Pipeline}

GENO features a modeling language and a solver that are tightly
coupled. The modeling language allows to specify a whole class of
optimization problems in terms of an objective function and
constraints that are given as vectorized linear algebra
expressions. Neither the objective function nor the constraints need
to be differentiable. Non-differentiable problems are transformed into
constrained, differentiable problems. A general purpose solver for
constrained, differentiable problems is then instantiated with the
objective function, the constraint functions and their respective
gradients. The gradients are computed by the matrix calculus
algorithm that has been recently published
in~\cite{LaueMG18}. The tight integration of the modeling language and
the solver is possible only because of this recent progress in
computing derivatives of vectorized linear algebra expressions.

Generating a solver takes only a few milliseconds. Once it has been
generated the solver can be used like any hand-written solver for
every instance of the specified problem class.  An online interface
to the GENO framework can be found at
\href{http://www.geno-project.org}{\texttt{http://www.geno-project.org}}.

\subsection{Modeling Language}

\begin{minipage}{0.68\textwidth}
  A GENO specification has four blocks (cf.\ the example to the right
  that shows an $\ell_1$-norm minimization problem from compressed
  sensing where the signal is known to be an element from the unit
  simplex.): (1) Declaration of the problem parameters that can be of
  type \emph{Matrix}, \emph{Vector}, or \emph{Scalar}, (2) declaration
  of the optimization variables that also can be of type
  \emph{Matrix}, \emph{Vector}, or \emph{Scalar}, (3) specification of
  the objective function in a MATLAB-like syntax, and finally (4)
  specification of the constraints, also in a MATLAB-like syntax that
  supports the following operators and functions: \texttt{+, -, *, /,
    .*, ./, $\wedge$, $.\wedge$, log, exp, sin, cos, tanh, abs, norm1,
    norm2, sum, tr, det, inv}. The set of operators and functions can
  be expanded when needed.
\end{minipage}
\quad
\begin{minipage}{0.28\textwidth}
  \begin{Verbatim}[frame=single]
 parameters
      Matrix A
      Vector b
 variables
      Vector x 
 min 
      norm1(x)
 st
      A*x == b
      sum(x) == 1
      x >= 0
  \end{Verbatim}
\end{minipage}

Note that in contrast to instance-based modeling languages like CVXPY
no dimensions have to be specified. Also, the specified problems do
not need to be convex. In the non-convex case, only a local optimal
solution will be computed.

\subsection{Generic Optimizer}

At its core, GENO's generic optimizer is a solver for unconstrained,
smooth optimization problems. This solver is then extended to handle
also non-smooth and constrained problems. In the following we first
describe the smooth, unconstrained solver before we detail how it is
extended to handling non-smooth and constrained optimization problems.

\paragraph{Solver for unconstrained, smooth problems.}

There exist quite a number of algorithms for unconstrained
optimization. Since in our approach we target problems with a few
dozen up to a few million variables, we decided to build on a
first-order method. This still leaves many options. Nesterov's
method~\cite{Nesterov83} has an optimal theoretical running time, that
is, its asymptotic running time matches the lower bounds in $\Omega
(1/\sqrt{\varepsilon})$ in the smooth, convex case and $\Omega(\log(1/
\varepsilon))$ in the strongly convex case with optimal dependence on
the Lipschitz constants $L$ and $\mu$ that have to be known in
advance. Here $L$ and $\mu$ are upper and lower bounds, respectively,
on the eigenvalues of the Hessian. On quadratic problems quasi-Newton
methods share the same optimal convergence
guarantee~\cite{huang70,Nazareth79} without requiring the values for
these parameters. In practice, quasi-Newton methods often outperform
Nesterov's method, although they cannot beat it in the worst case. It
is important to keep in mind that theoretical running time guarantees
do not always translate into good performance in practice. For
instance, even the simple subgradient method has been shown to have a
convergence guarantee in $O(\log(1/ \varepsilon))$ on strongly convex
problems~\cite{Goffin77}, but it is certainly not competitive on real
world problems.

Hence, we settled on a quasi-Newton method and implemented the
well-established \mbox{L-BFGS-B} algorithm~\cite{ByrdLNZ95,ZhuBLN97}
that can also handle box constraints on the variables. It serves as
the solver for unconstrained, smooth problems. The algorithm combines
the standard limited memory quasi-Newton method with a projected
gradient path approach. In each iteration, the gradient path is
projected onto the box constraints and the quadratic function based on
the second-order approximation (L-BFGS) of the Hessian is minimized
along this path. All variables that are at their boundaries are fixed
and only the remaining free variables are optimized using the
second-order approximation. Any solution that is not within the bound
constraints is projected back onto the feasible set by a simple
min/max operation~\cite{MoralesN11}. Only in rare cases, a projected
point does not form a descent direction. In this case, instead of
using the projected point, one picks the best point that is still
feasible along the ray towards the solution of the quadratic
approximation. Then, a line search is performed for satisfying the
strong Wolfe conditions~\cite{Wolfe69,Wolfe71}. This condition is
necessary for ensuring convergence also in the non-convex case. The
line search also obliterates the need for a step length or learning
rate that is usually necessary in SGD, subgradient algorithms, or
Nesterov's method. Here, we use the line search proposed
in~\cite{MoreT94} which we enhanced by a simple backtracking line
search in case the solver enters a region where the function is not
defined.

\paragraph{Solver for unconstrained non-smooth problems.}

Machine learning often entails non-smooth optimization problems, for
instance all problems that employ $\ell_1$-regularization. Proximal
gradient methods are a general technique for addressing such
problems~\cite{Proximal15}. Here, we pursue a different approach.  All
non-smooth convex optimization problems that are allowed by our
modeling language can be written as $\min_x \{\max_i f_i(x)\}$ with
smooth functions $f_i(x)$~\cite{Nesterov05}.  This class is flexible
enough to accommodate most of the non-smooth objective functions
encountered in machine learning. All problems in this class can be
transformed into constrained, smooth problems of the form
\[
\begin{array}{rl}  \displaystyle
\min_{t, x} &  t \\
\st&  f_i(x) \leq t.
\end{array}
\]
The transformed problems can then be solved by the solver for
constrained, smooth optimization problems that we describe next.

\paragraph{Solver for smooth constrained problems.}

There also quite a few options for solving smooth, constrained
problems, among them projected gradient methods, the alternating
direction method of multipliers
(ADMM)~\cite{boydADMM,Gabay1976,Glowinski1975}, and the augmented
Lagrangian approach~\cite{Hestenes69,Powell69}. For GENO, we decided
to follow the augmented Lagrangian approach, because this allows us to
(re-)use our solver for smooth, unconstrained problems directly. Also,
the augmented Lagrangian approach is more generic than ADMM. All
ADMM-type methods need a proximal operator that cannot be derived
automatically from the problem specification and a closed-form
solution is sometimes not easy to compute. Typically, one uses
standard duality theory for deriving the
prox-operator. In~\cite{Proximal15}, prox-operators are tabulated for
several functions.

The augmented Lagrangian method can be used for solving the following
general standard form of an abstract constrained optimization problem
\begin{equation} \label{eq:constrained}
\begin{array}{rl}  \displaystyle
  \min_{x} &  f(x) \\
  \st &  h(x) = 0 \\
  	& g(x) \leq 0,
  	\end{array}
\end{equation}
where $x\in\setR^n$, $f\colon \setR^n \to \setR$,
$h\colon\setR^n\to\setR^m$, $g\colon\setR^n\to\setR^p$ are
differentiable functions, and the equality and inequality constraints
are understood component-wise.

The augmented Lagrangian of Problem~\eqref{eq:constrained} is the
following function
\[ 
  L_\rho (x, \lambda, \mu) = f(x) + \frac{\rho}{2}
  \norm{h(x)+\frac{\lambda}{\rho}}^2 + \frac{\rho}{2} \norm{\left(g(x)
    + \frac{\mu}{\rho}\right)_+}^2,
\]
where $\lambda\in\setR^m$ and $\mu\in\setR_{\geq 0}^p$ are Lagrange
multipliers, $\rho >0$ is a constant, $\norm{\cdot}$ denotes the
Euclidean norm, and $(v)_+$ denotes $\max\{v, 0\}$. The Lagrange
multipliers are also referred to as dual variables. In principle, the
augmented Lagrangian is the standard Lagrangian of
Problem~\eqref{eq:constrained} augmented with a quadratic penalty
term. This term provides increased stability during the optimization
process which can be seen for example in the case that
Problem~\eqref{eq:constrained} is a linear program. Note, that
whenever Problem~\eqref{eq:constrained} is convex, i.e., $h$ are
affine functions and $g$ are convex in each component, then the
augmented Lagrangian is also a convex function.

The Augmented Lagrangian Algorithm~\ref{algo:1} runs in iterations.
In each iteration it solves an unconstrained smooth optimization
problem. Upon convergence, it will return an approximate solution $x$
to the original problem along with an approximate solution of the
Lagrange multipliers for the dual problem. If
Problem~\eqref{eq:constrained} is convex, then the algorithm returns
the global optimal solution. Otherwise, it returns a local
optimum~\cite{Bertsekas99}. The update of the multiplier $\rho$ can be
ignored and the algorithm still converges~\cite{Bertsekas99}. However,
in practice it is beneficial to increase it depending on the progress
in satisfying the constraints~\cite{Birgin14}. If the infinity norm of
the constraint violation decreases by a factor less than $\tau=1/2$ in
one iteration, then $\rho$ is multiplied by a factor of two.
\begin{algorithm}[h!]
  \caption{Augmented Lagrangian Algorithm}
  \label{algo:1}
  \begin{algorithmic}[1]
    \STATE {\bfseries input:} instance of Problem~\ref{eq:constrained}
    \STATE {\bfseries output:} approximate solution $x\in\setR^{n},
    \lambda\in\setR^{p}, \mu\in\setR_{\geq 0}^{m}$ 
    \STATE initialize $x^0 = 0$, $\lambda^0 = 0$, $\mu^0 = 0$, and $\rho=1$
    \REPEAT
    \STATE  $x^{k+1} :=\quad \argmin_{x}\, L_\rho(x, \lambda^k, \mu^k)$ \label{algo:x}
    \STATE $\lambda^{k+1} :=\quad  \lambda^k + \rho h(x^{k+1})$ \label{algo:lambda}
    \STATE $\mu^{k+1} :=\quad  \left(\mu^k + \rho g(x^{k+1})\right)_+$ \label{algo:mu}
    \STATE update $\rho$ \label{algo:rho}
    \UNTIL{convergence}
    \RETURN $x^k, \lambda^k, \mu^k$
  \end{algorithmic}
\end{algorithm}

\section{Limitations}

While GENO is very general and efficient, as we will demonstrate in
the experimental Section~\ref{sec:experiments}, it also has some
limitations that we discuss here.  For small problems, i.e., problems
with only a few dozen variables, Newton-type methods with a direct
solver for the Newton system can be even faster.  GENO also does not
target deep learning applications, where gradients do not need to be
computed fully but can be sampled.

Some problems can pose numerical problems, for instance problems
containing an $\exp$ operator might cause an
overflow/underflow. However, this is a problem that is faced by all
frameworks. It is usually addressed by introducing special operators
like \emph{logsumexp}.

Furthermore, GENO does not perform sanity checks on the provided
input. Any syntactically correct problem specification is accepted by
GENO as a valid input. For example, $\log(\det(xx^\top))$, where $x$
is a vector, is a valid expression. But the determinant of the outer
product will always be zero and hence, taking the logarithm will
fail. It lies within the responsibility of the user to make sure that
expressions are mathematically valid.

\section{Experiments} \label{sec:experiments}

We conducted a number of experiments to show the wide applicability
and efficiency of our approach. For the experiments we have chosen
classical problems that come with established well-engineered solvers
like logistic regression or elastic net regression, but also problems
and algorithms that have been published at NeurIPS and ICML only
within the last few years. The experiments cover smooth unconstrained
problems as well as constrained, and non-smooth problems. To prevent a
bias towards GENO, we always used the original code for the competing
methods and followed the experimental setup in the papers where these
methods have been introduced. We ran the experiments on standard data
sets from the LIBSVM data set repository, and, in some cases, on
synthetic data sets on which competing methods had been evaluated in
the corresponding papers.

Specifically, our experiments cover the following problems and
solvers: $\ell_1$- and $\ell_2$-regularized logistic regression,
support vector machines, elastic net regression, non-negative least
squares, symmetric non-negative matrix factorization, problems from
non-convex optimization, and compressed sensing. Among other
algorithms, we compared against a trust-region Newton method with
conjugate gradient descent for solving the Newton system, sequential
minimal optimization (SMO), dual coordinate descent, proximal methods
including ADMM and variants thereof, interior point methods,
accelerated and variance reduced variants of SGD, and Nesterov's
optimal gradient descent. Please refer to the appendix
for more details on the solvers and GENO models.

Our test machine was equipped with an eight-core Intel Xeon
CPU~E5-2643 and 256GB RAM. As software environment we used Python~3.6,
along with NumPy~1.16, SciPy~1.2, and scikit-learn~0.20. In some cases
the original code of competing methods was written and run in
MATLAB~R2019. The solvers generated by GENO spent between $80\%$ and
$99\%$ of their time on evaluating function values and
gradients. Here, these evaluations essentially reduce to evaluating
linear algebra expressions. Since all libraries are linked against the
Intel MKL, running times of the GENO solvers are essentially the same
in both environments, Python and MATLAB, respectively.

\subsection{Regularization Path for $\ell_1$-regularized Logistic Regression}

\begin{figure*}[t!]
  \centering
  \begin{tabular}{ccc}
    \includegraphics[width=0.48\textwidth]{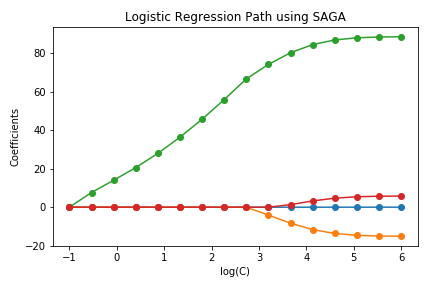} &
    \includegraphics[width=0.48\textwidth]{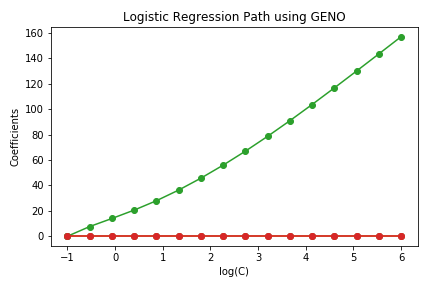} \\
    \includegraphics[width=0.48\textwidth]{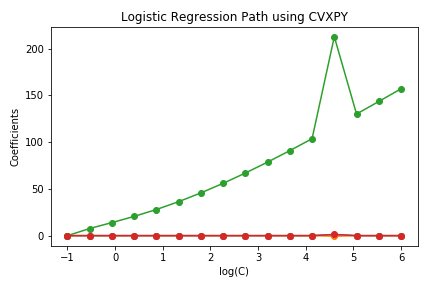} &
    \includegraphics[width=0.48\textwidth]{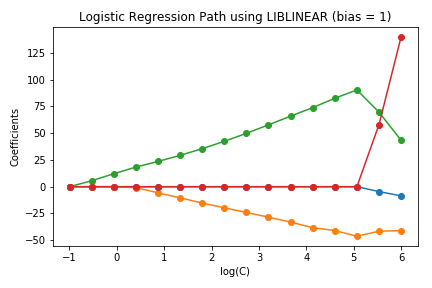} &
  \end{tabular}
  \caption{The regularization path of $\ell_1$-regularized logistic
    regression for the Iris data set using SAGA, GENO, CVXPY, and
    LIBLINEAR.}
  \label{fig:logRegL1}
\end{figure*}

Computing the regularization path of the $\ell_1$-regularized logistic
regression problem~\cite{Cox58} is a classical machine learning
problem, and only boring at a first glance. The problem is well suited
for demonstrating the importance of both aspects of our approach,
namely flexibility and efficiency. As a standard problem it is covered
in scikit-learn. The scikit-learn implementation features the SAGA
algorithm~\cite{DefazioBL14} for computing the whole regularization
path that is shown in Figure~\ref{fig:logRegL1}. This figure can also
be found on the scikit-learn
website~\footnote{\href{{https://scikit-learn.org/stable/auto_examples/linear_model/plot_logistic_path.html}}{https://scikit-learn.org/stable/auto\_examples/linear\_model/plot\_logistic\_path.html}}. However,
when using GENO, the regularization path looks different, see also
Figure~\ref{fig:logRegL1}. Checking the objective functions values
reveals that the precision of the SAGA algorithm is not enough for
tracking the path faithfully. GENO's result can be reproduced by using
CVXPY except for one outlier at which CVXPY did not compute the
optimal solution. LIBLINEAR~\cite{FanCHWL08,ZhuangJYL18} can also be
used for computing the regularization path, but also fails to follow
the exact path. This can be explained as follows: LIBLINEAR also does
not compute optimal solutions, but more importantly, in contrast to
the original formulation, it penalizes the bias for algorithmic
reasons. Thus, changing the problem slightly can lead to fairly
different results.

CVXPY, like GENO, is flexible and precise enough to accommodate the
original problem formulation and to closely track the regularization
path. But it is not as efficient as GENO. On the problem used in
Figure~\ref{fig:logRegL1} SAGA takes 4.3 seconds, the GENO solver
takes 0.5 seconds, CVXPY takes 13.5 seconds, and LIBLINEAR takes 0.05
seconds but for a slightly different problem and insufficient
accuracy.

\subsection{$\ell_2$-regularized Logistic Regression}

Logistic regression is probably the most popular linear, binary
classification method. It is given by the following unconstrained
optimization problem with a smooth objective function
\[
\begin{array}{rl} \displaystyle
  \min_{w} & \frac{\lambda}{2} \norm{w}_2^2 + \frac{1}{m} \sum_i
  \log(\exp(-y_i X_i w) + 1),
\end{array}
\]
where $X\in\setR^{m\times n}$ is a data matrix, $y\in\{-1, +1\}^m$ is
a label vector, and $\lambda\in\setR$ is the regularization parameter.
Since it is a classical problem there exist many well-engineered
solvers for $\ell_2$-regularized logistic regression. The problem also
serves as a testbed for new algorithms. We compared GENO to the
parallel version of LIBLINEAR and a number of recently developed
algorithms and their implementations, namely
Point-SAGA~\cite{Defazio16}, SDCA~\cite{Shalev-Shwartz13}, and
catalyst SDCA~\cite{LinMH15}). The latter algorithms implement some
form of SGD. Thus their running time heavily depends on the values for
the learning rate (step size) and the momentum parameter in the case
of accelerated SGD. The best parameter setting often depends on the
regularization parameter and the data set. We have used the code
provided by~\cite{Defazio16} and the parameter settings therein.

\begin{figure*}[h!]
  \centering
  \begin{tabular}{ccc}
    \includegraphics[width=0.3\textwidth]{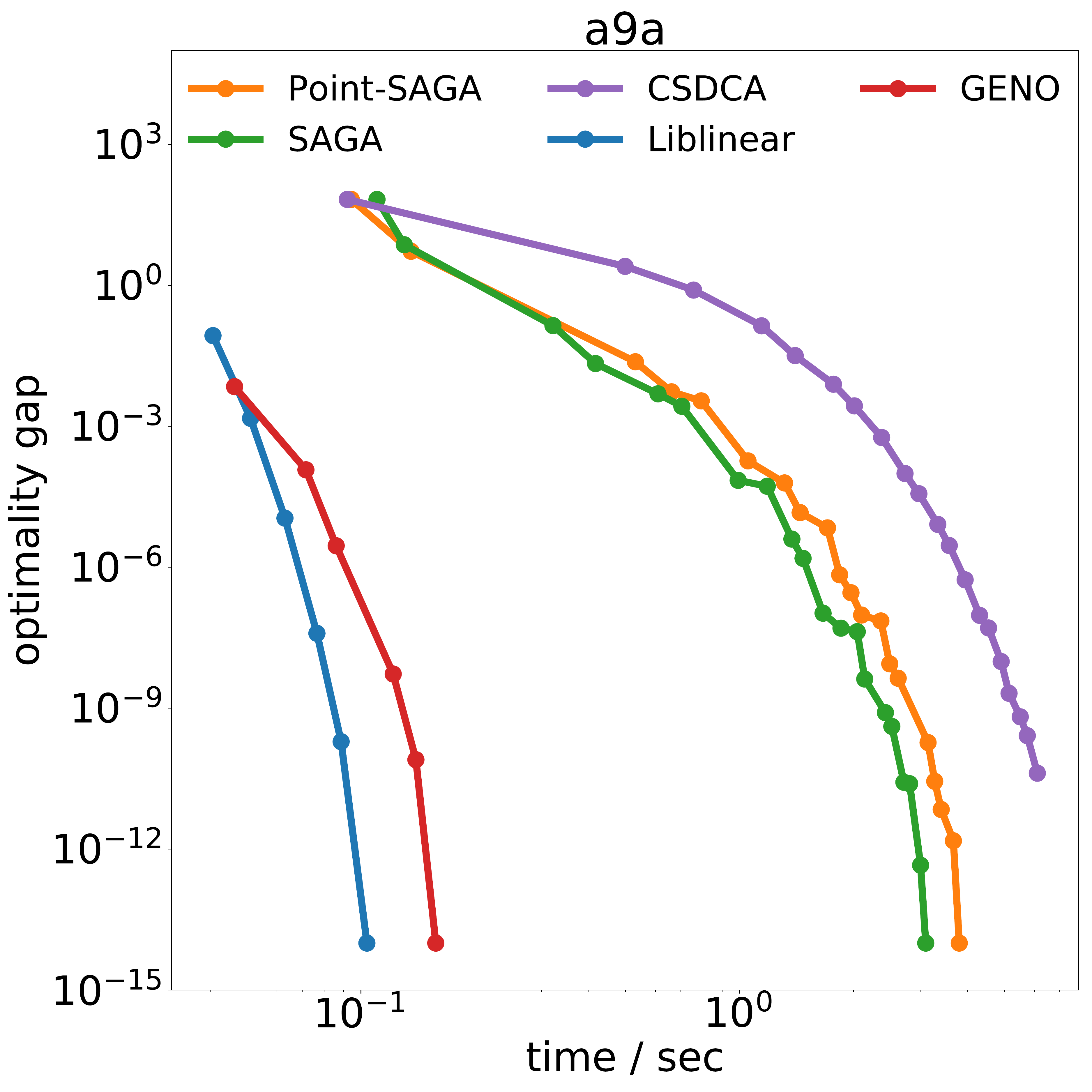} &
    \includegraphics[width=0.3\textwidth]{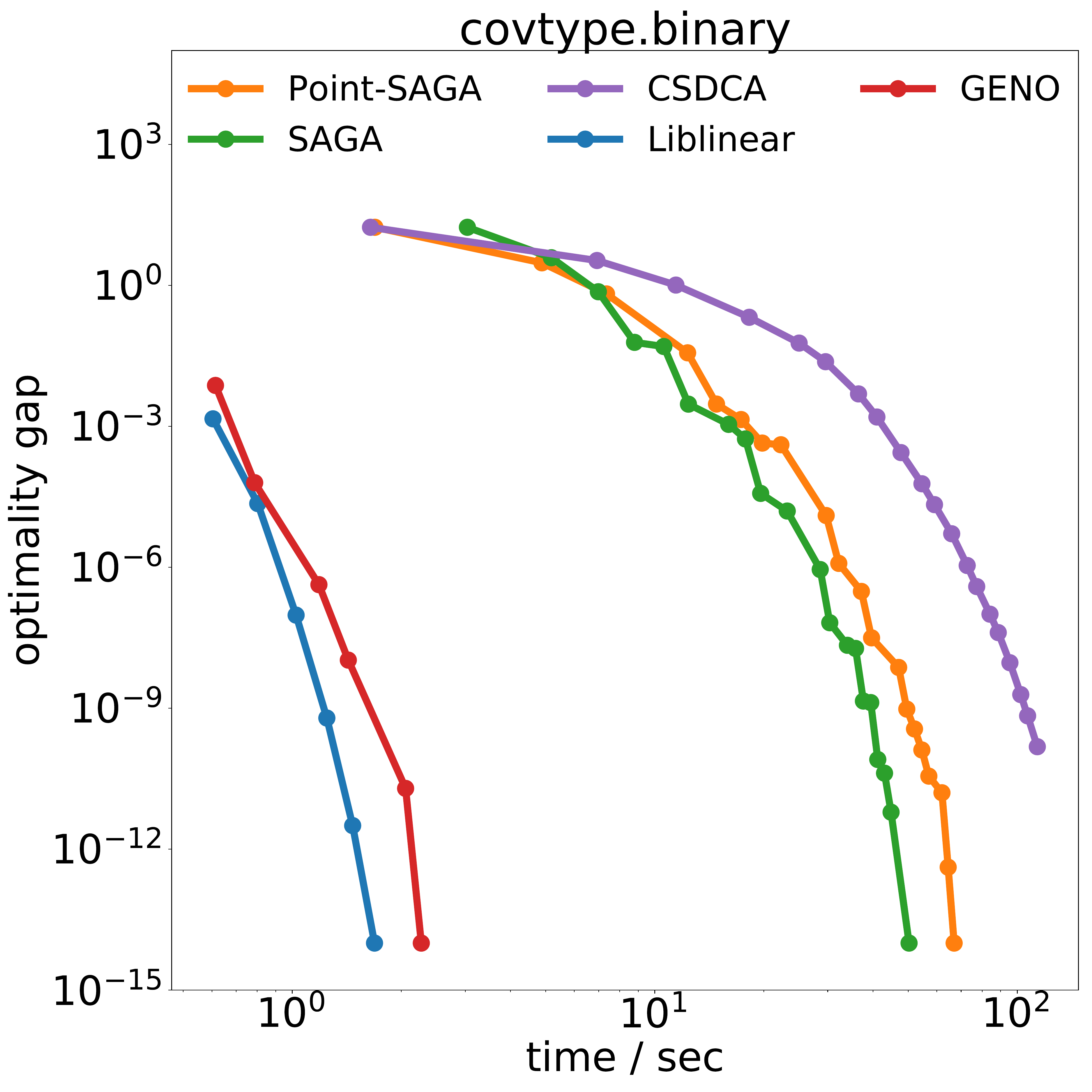} &
    \includegraphics[width=0.3\textwidth]{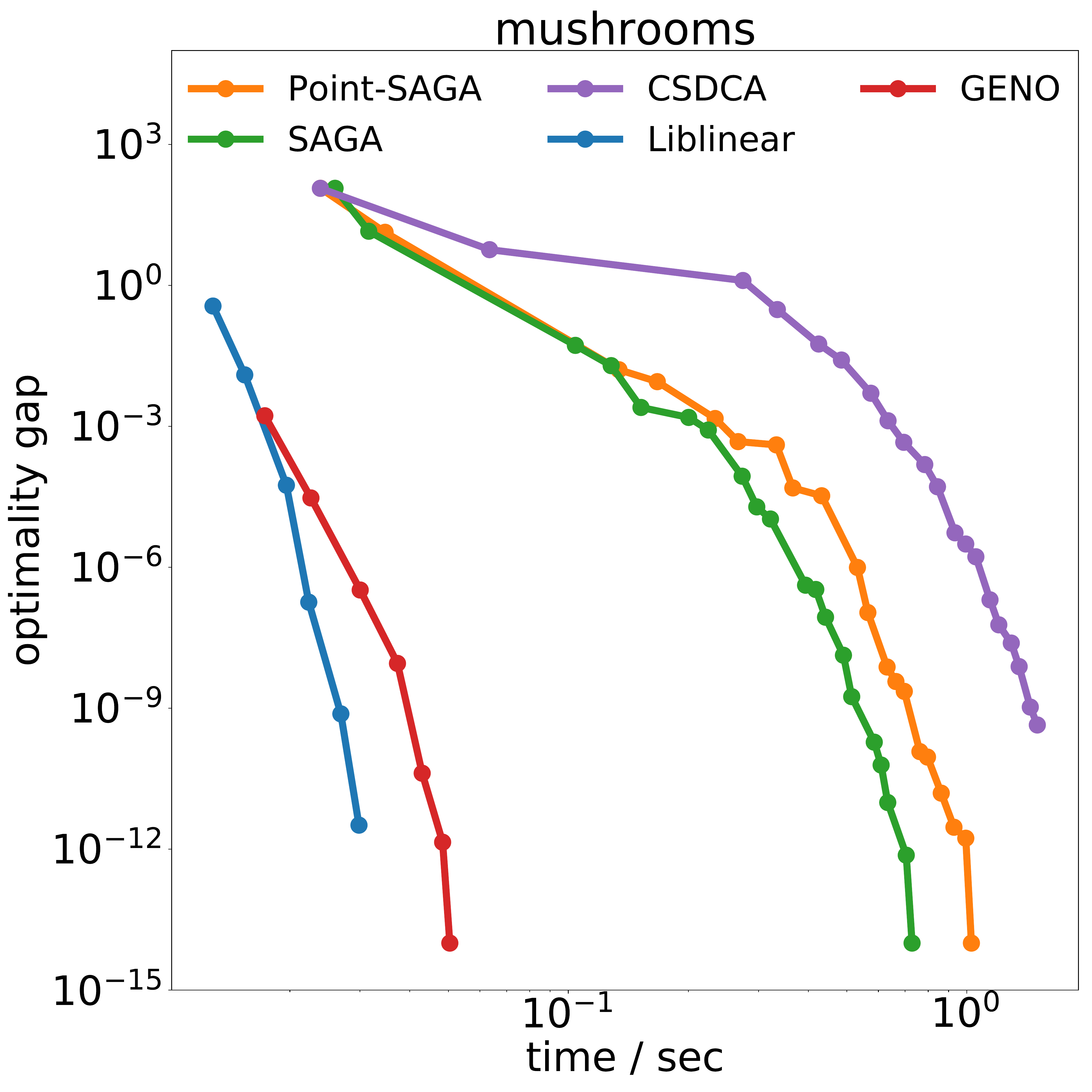} \\
    \\
    \includegraphics[width=0.3\textwidth]{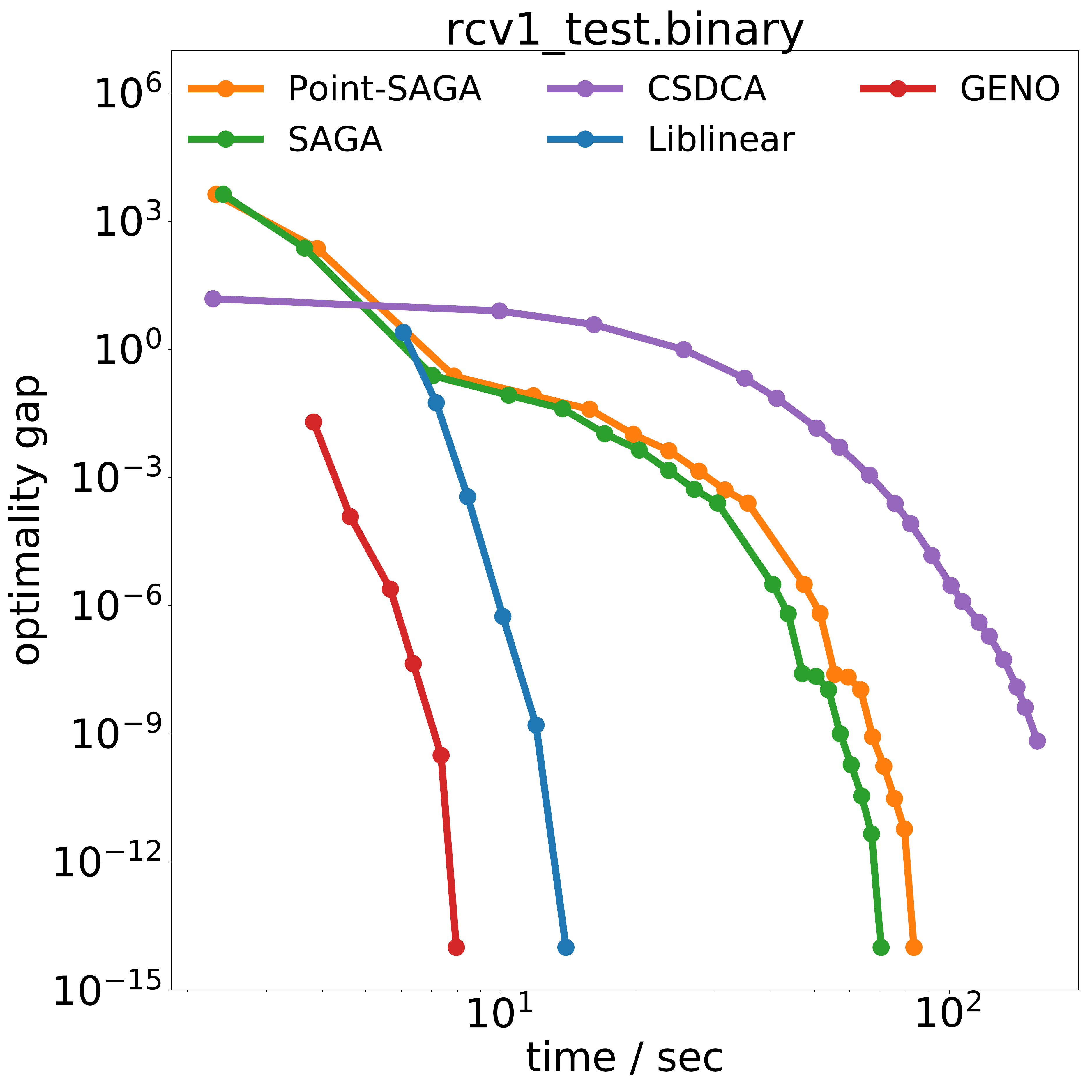}  &
    \includegraphics[width=0.3\textwidth]{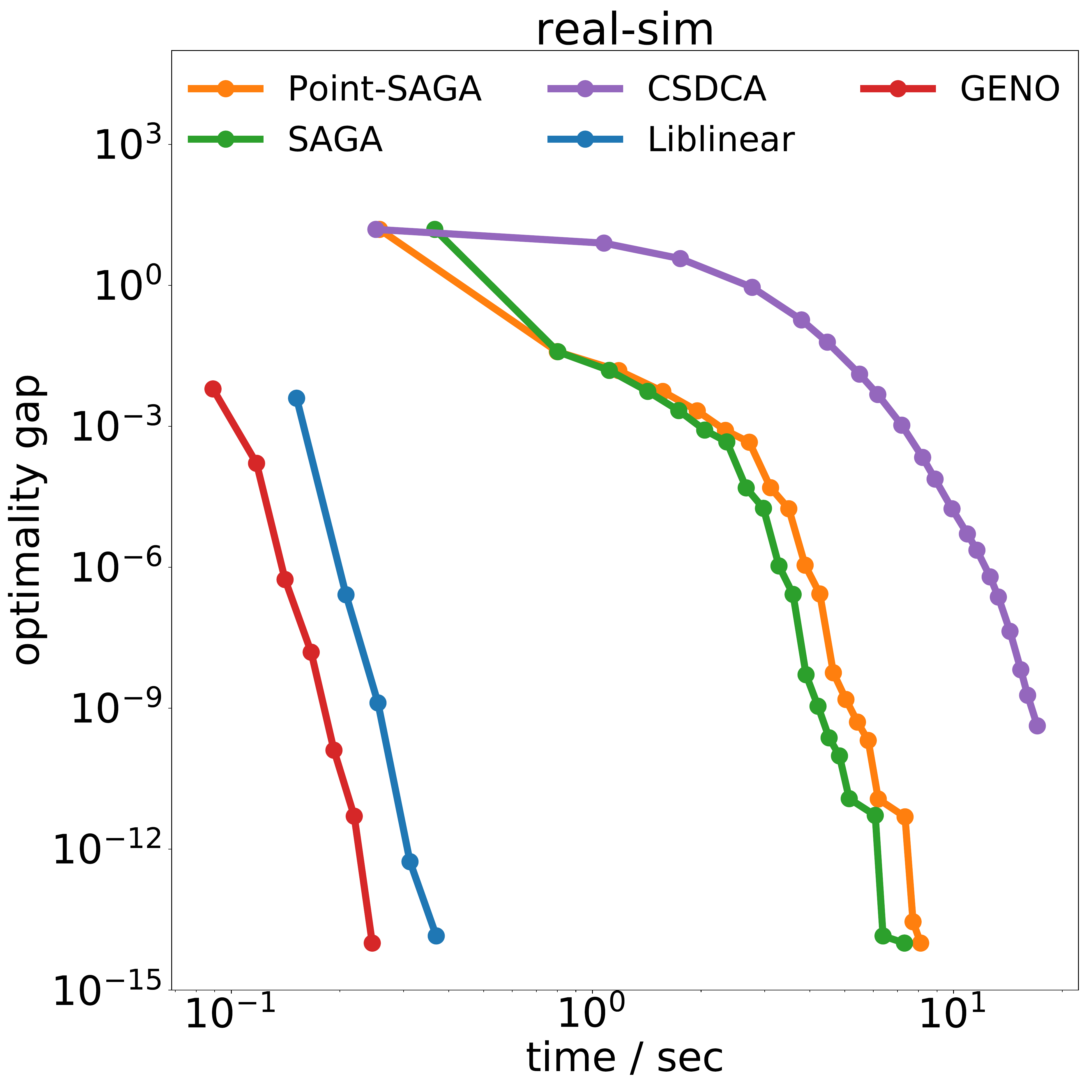} &
    \includegraphics[width=0.3\textwidth]{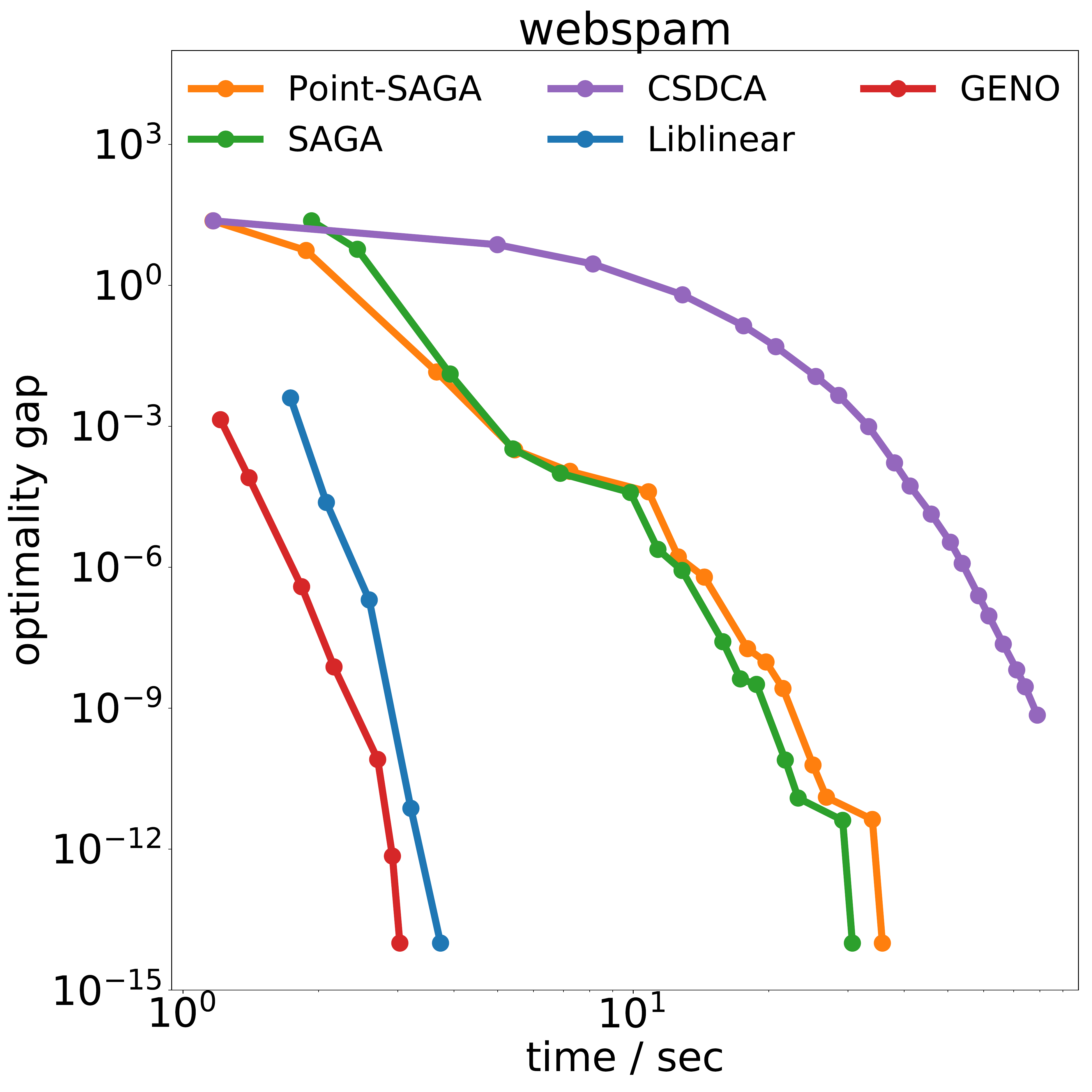}
  \end{tabular}
  \caption{Running times for different solvers on the
    $\ell_2$-regularized logistic regression problem.}
  \label{fig:lrl2}
\end{figure*}

For our experiments we set the regularization parameter
$\lambda=10^{-4}$ and used real world data sets that are commonly used
in experiments involving logistic regression. GENO converges as
rapidly as LIBLINEAR and outperforms any of the recently published
solvers by a good margin, see Figure~\ref{fig:lrl2}.

On substantially smaller data sets we also compared GENO to CVXPY with
both the ECOS~\cite{Ecos1} and the SCS solver~\cite{scs}.
As can be seen from Table~\ref{tab:lrl2}, GENO is orders of magnitude
faster. 
\begin{table}[h!]
  \centering
  \caption{Running times in seconds for different general purpose
    solvers on small instances of the $\ell_2$-regularized logistic
    regression problem. The approximation error is close to $10^{-6}$
    for all solvers.}
  \label{tab:lrl2}
  \begin{tabular}{l*{8}{r}}
    \toprule
    \multirow{2}{*}{Solver} & \multicolumn{7}{c}{Data sets} \\
    \cmidrule{2-8}
    & heart & ionosphere & breast-cancer & australian & diabetes & a1a & a5a \\
    \midrule
    GENO & 0.005 & 0.013 & 0.004 & 0.014 & 0.006 & 0.023  & 0.062\\
    ECOS & 1.999 & 2.775 & 5.080 & 5.380 & 5.881 & 12.606 & 57.467\\
    SCS  & 2.589 & 3.330 & 6.224 & 6.578 & 6.743 & 16.361 & 87.904\\
    \bottomrule
  \end{tabular}
\end{table}

\subsection{Support Vector Machines}

Support Vector Machines (SVMs)~\cite{CortesV95} have been studied
intensively and are widely used, especially in combination with
kernels~\cite{SchoelkopfS01}. They remain populat, as is indicated by
the still rising citation count of the popular and heavily-cited
solver LIBSVM~\cite{ChangL01}. The dual formulation of an SVM is given as the
following quadratic optimization problem
\[
\begin{array}{rl}
  \min_{a} &   \frac{1}{2}(a\odot y)^\top K (a\odot y) - \norm{a}_1\\
  \st & y^\top a = 0\\
	  & 0 \leq a\leq c,
\end{array}
\]
where $K\in\setR^{m\times m}$ is a kernel matrix, $y\in\{-1, +1\}^m$
is a binary label vector, $c\in\setR$ is the regularization parameter,
and $\odot$ is the element-wise multiplication. While the SVM problem
with a kernel can also be solved in the primal~\cite{chapelle2007}, it
is traditionally solved in the dual.  We use a Gaussian kernel, i.e.,
$K_{ij} =\exp\left(-\gamma\norm{X_i - X_j}_2^2\right)$ and standard
data sets. We set the bandwith parameter $\gamma=1/2$ which
corresponds to roughly the median of the pairwise data point distances
and set $C=1$. Table~\ref{tab:svm} shows that the solver generated by
GENO is as efficient as LIBSVM which has been maintained and improved
over the last 15 years. Both solvers outperform general purpose
approaches like CVXPY with OSQP~\cite{osqp}, SCS~\cite{scs},
Gurobi~\cite{gurobi}, or Mosek~\cite{mosek} by a few orders of
magnitude.
\begin{table}[h]
  \centering
  \caption{Running times in seconds for solving a dual
    Gaussian-kernelized SVM. The optimality gap is close to $10^{-4}$
    for all solvers and data sets. Missing entries in the table
    indicate that the solver did not finish within one hour. }
  \label{tab:svm}
  \begin{tabular}{l*{8}{r}}
    \toprule
    \multirow{2}{*}{Solver} & \multicolumn{8}{c}{Datasets} \\
    \cmidrule{2-9}
    & ionosphere & australian & diabetes & a1a & a5a & a9a & w8a &  cod-rna \\
    \midrule
    GENO & 0.009 & 0.024 & 0.039 & 0.078 & 1.6 & 30.0 & 25.7  & 102.1\\
    LIBSVM & 0.005 & 0.010 & 0.009 & 0.088 & 1.0 & 18.0 & 78.6 & 193.1\\
    SCS  & 0.442 & 1.461 & 3.416 & 11.707 & 517.5 &  -& - & -\\
    OSQP & 0.115 & 0.425 & 0.644 & 3.384 & 168.2 & - & - & -\\
    Gurobi & 0.234 & 0.768 & 0.992 & 4.307 & 184.4 & - & - &- \\
    Mosek & 0.378 & 0.957 & 1.213 & 6.254 & 152.7 & - & - & - \\
    \bottomrule
  \end{tabular}
\end{table}

\subsection{Elastic Net}

Elastic net regression~\cite{ZouH05} has also been studied intensively
and is used mainly for mircoarray data classification and gene
selection. Given some data $X\in\setR^{m\times n}$ and a response
$y\in\setR^{m}$, elastic net regression seeks to minimize
\[
\frac{1}{2m}\norm{Xw-y}^2_2 + \alpha\left( \lambda\norm{w} + \frac{1-\lambda}{2}\norm{w}_2^2\right),
\]
where $\alpha$ and $\lambda$ are the corresponding elastic net
regularization parameters.  The most popular solver is glmnet, a dual
coordinate descent approach that has been implementated in
Fortran~\cite{FriedmanHT09}. In our experiments, we follow the same
setup as in~\cite{FriedmanHT09}. We generated Gaussian data
$X\in\setR^{m\times n}$ with $m$ data points and $n$ features. The
outcome values $y$ were generated by
\[
y = \sum_{j=1}^n X_j \beta_j + k\cdot z,
\] 
where $\beta_j = (-1)^j \exp(-j/10)$, $z\sim {\cal{N}} (0, 1)$, and
$k$ is chosen such that the signal-to-noise ratio is 3. We varied the
number of data points $m$ and the number of features $n$. The results
are shown in Table~\ref{tab:elasticnet}.  It can be seen that the
solver generated by GENO is as efficient as glment and orders of
magnitude faster than comparable state-of-the-art general purpose
approaches like CVXPY coupled with ECOS, SCS, Gurobi, or Mosek. Note,
that the OSQP solver could not be run on this problem since CVXPY
raised the error that it cannot convert this problem into a QP.
\begin{table}[h]
  \centering
  \caption{Running times for the elastic net regression problem in
    seconds. Missing entries in the table indicate that the solver did
    not finish within one hour. The optimality gap is about $10^{-8}$
    for all solvers which is the standard setting for glmnet.}
  \label{tab:elasticnet}
  \begin{tabular}{r*{8}{r}}
    \toprule
    \multirow{2}{*}{m} & \multirow{2}{*}{n} & \multicolumn{6}{c}{Solvers} \\
    \cmidrule{3-8}
    & & GENO & glmnet & ECOS & SCS & Gurobi & Mosek \\
    \midrule
    1000 & 1000 & 0.11 & 0.10 &   43.27 &  2.33 &  21.14 & 1.77 \\
    2000 & 1000 & 0.14 & 0.08 &  202.04 &  9.24 &  58.44 & 3.52 \\
    3000 & 1000 & 0.18 & 0.08 &  513.78 & 22.86 & 114.79 & 5.38 \\
    4000 & 1000 & 0.21 & 0.09 &  -      & 38.90 & 185.79 & 7.15 \\
    5000 & 1000 & 0.27 & 0.11 &  -      & 13.88 & 151.08 & 8.69 \\
    \midrule
    1000 & 5000 & 1.74 & 0.62 & - &  28.69 & - & 13.06 \\
    2000 & 5000 & 1.49 & 1.41 & - &  45.79 & - & 27.69 \\
    3000 & 5000 & 1.58 & 2.02 & - &  81.83 & - & 50.99 \\
    4000 & 5000 & 1.24 & 1.88 & - & 135.94 & - & 67.60 \\
    5000 & 5000 & 1.41 & 1.99 & - & 166.60 & - & 71.92 \\
    \midrule
    5000   & 10000 &  4.11 &  4.75 & - & - & - & - \\
    7000   & 10000 &  4.76 &  5.52 & - & - & - & - \\
    10000  & 10000 &  4.66 &  3.89 & - & - & - & - \\
    50000  & 10000 & 13.97 &  6.34 & - & - & - & - \\
    70000  & 10000 & 18.82 & 11.76 & - & - & - & - \\
    100000 & 10000 & 23.38 & 23.42 & - & - & - & - \\
    \bottomrule
  \end{tabular}
\end{table}

\subsection{Non-negative Least Squares}

\begin{figure*}[h]
  \centering
  \begin{tabular}{ccc}
    \includegraphics[width=0.4\textwidth]{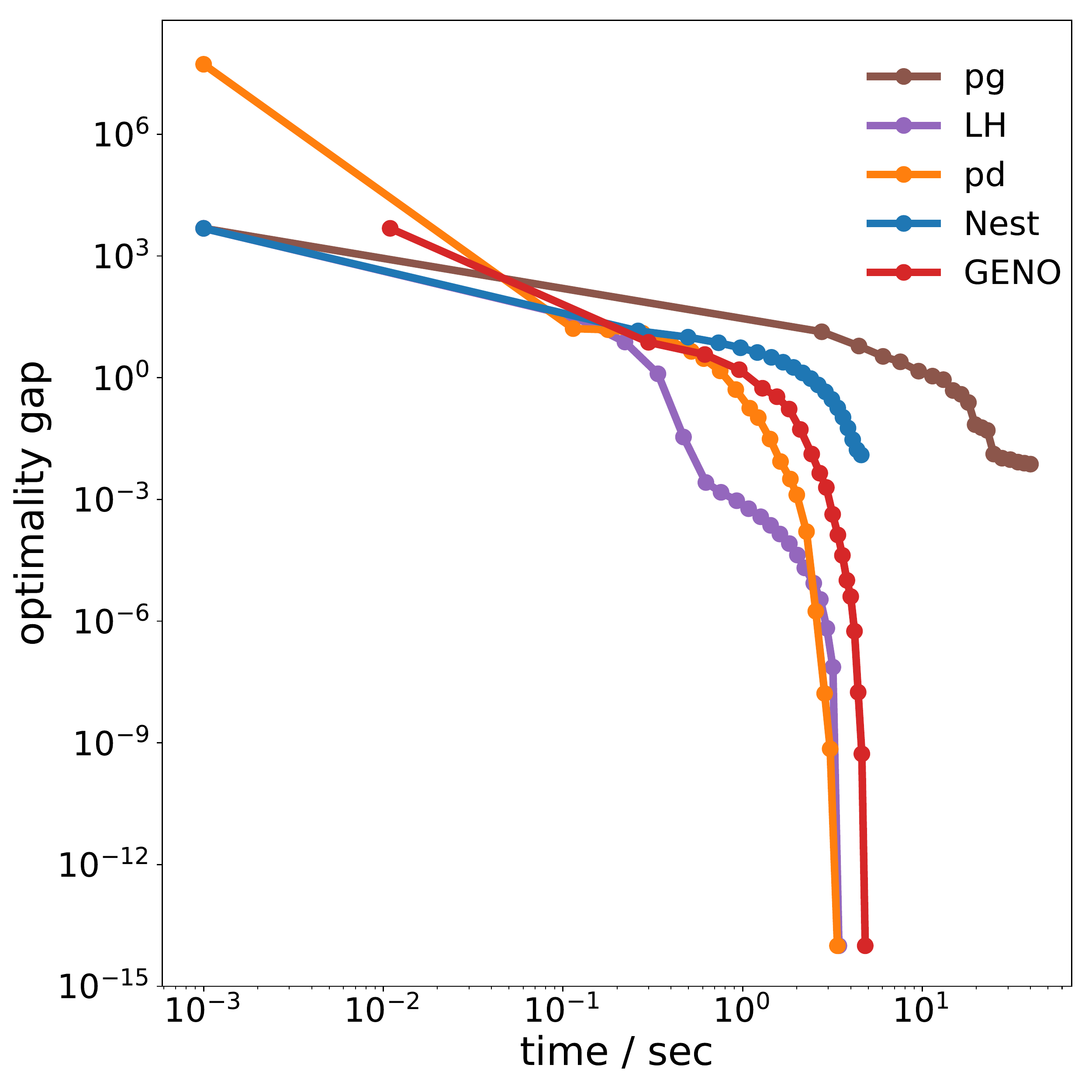} &
    \quad\quad & 
    \includegraphics[width=0.4\textwidth]{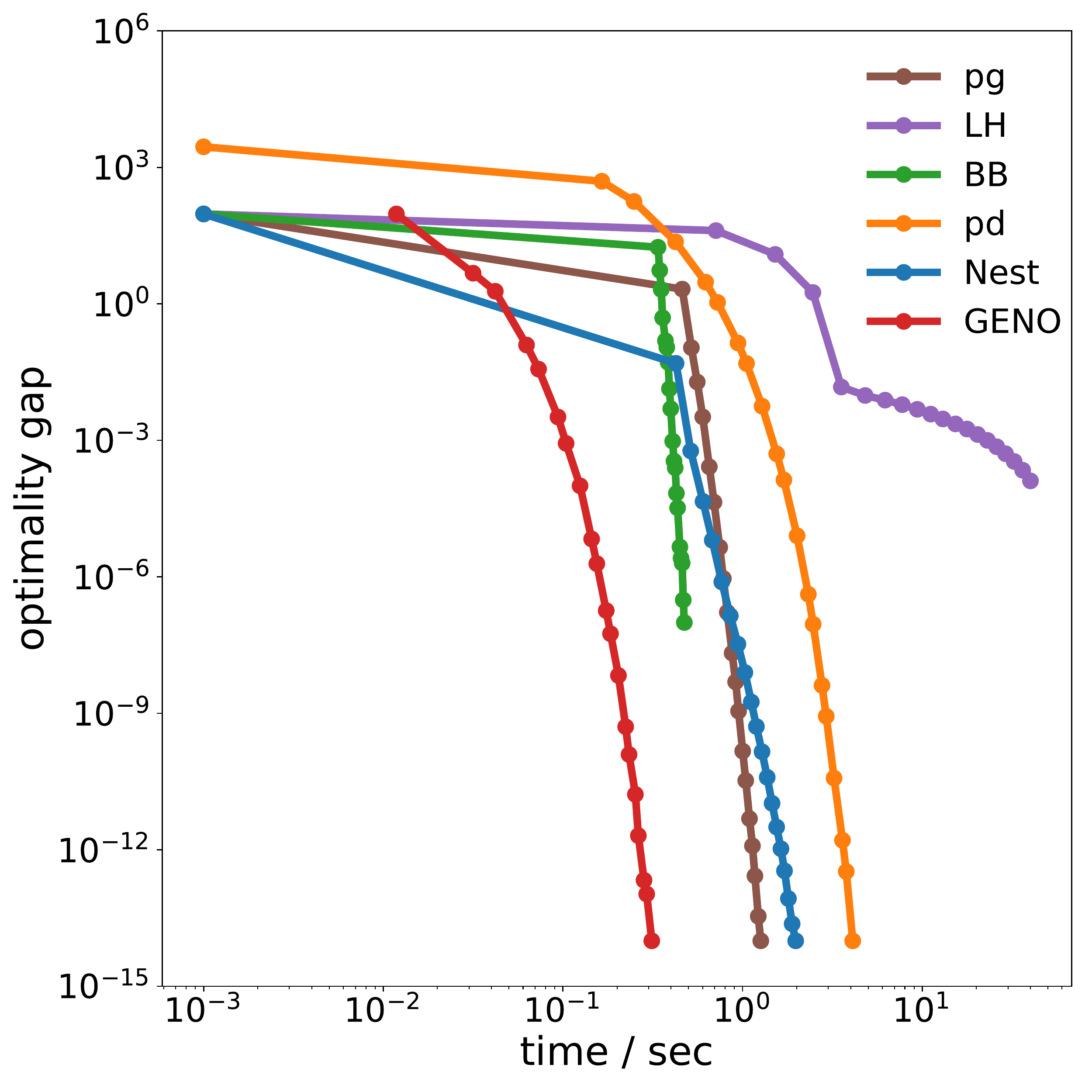} 
  \end{tabular}
  \caption{Running times for non-negative least squares
    regression. The figure on the left shows the running times for the
    experiment (i) and the figure on the right the running times for
    experiment (ii). The algorithms are projected gradient descent
    (pd), Lawson-Hanson (LH), subspace Barzilai-Borwein (BB),
    primal-dual interior point method (pd), Nesterov's accelerated
    projected gradient descent (Nest), and GENO.}
  \label{fig:nnls}
\end{figure*}

Least squares is probably the most widely used regression method.
Non-negative least squares is an extension that requires the output to
be non-negative. It is given as the following optimization problem
\[
\begin{array}{rl}
  \min_{x} & \norm{Ax-b}_2^2 \\
  \st & x\geq 0,
\end{array}
\]
where $A\in\setR^{m\times n}$ is a given design matrix and
$b\in\setR^m$ is the response vector.  Since non-negative least
squares has been studied intensively, there is a plenitude of solvers
available that implement different optimization methods. An overview
and comparison of the different methods can be found
in~\cite{Slawski13}. Here, we use the accompanying code described
in~\cite{Slawski13} for our comparison. We ran two sets of
experiments, similarly to the comparisons in~\cite{Slawski13}, where
it was shown that the different algorithms behave quite differently on
these problems. For experiment (i), we generated random data
$A\in\setR^{2000\times 6000}$, where the entries of $A$ were sampled
uniformly at random from the interval $[0, 1]$ and a sparse vector
$x\in\setR^{6000}$ with non-zero entries sampled also from the uniform
distribution of $[0, 1]$ and a sparsity of $0.01$. The outcome values
were then generated by $y = \sqrt{0.003}\cdot Ax + 0.003 \cdot z$,
where $z\sim {\cal{N}}(0, 1)$. For experiment (ii),
$A\in\setR^{6000\times 3000}$ was drawn form a Gaussian distribution
and $x$ had a sparsity of $0.1$. The outcome variable was generated by
$y=\sqrt{1/6000}\cdot Ax+0.003\cdot z$, where $z\sim {\cal{N}}(0,
1)$. The differences between the two experiments are the following:
(1) The Gram matrix $A^\top A$ is singular in experiment (i) and
regular in experiment (ii), (2) The design matrix $A$ has isotropic
rows in experiment (ii) which does not hold for experiment (i), and
(3) $x$ is significantly sparser in (i) than in (ii).  We compared the
solver generated by GENO with the following approaches: the classical
Lawson-Hanson algorithm~\cite{lawson95}, which employs an active set
strategy, a projected gradient descent algorithm combined with an
Armijo-along-projection-arc line search~\cite[Ch~2.3]{Bertsekas99}, a
primal-dual interior point algorithm that uses a conjugate gradient
descent algorithm~\cite{BoydV04} with a diagonal preconditioner for
solving the Newton system, a subspace Barzilai-Borwein
approach~\cite{kim13}, and Nesterov's accelerated projected gradient
descent~\cite{Nesterov83}. Figure~\ref{fig:nnls} shows the results for
both experiments. Note, that the Barzilai-Borwein approach with
standard parameter settings diverged on experiment (i) and it would
stop making progress on experiment (ii).  While the other approaches
vary in running time depending on the problem, the experiments show
that the solver generated by GENO is always among the fastest compared
to the other approaches.

We provide the final running times of the general purpose solvers in
Table~\ref{tab:nnls} since obtaining intermediate solutions is not
possible for these solvers.  Table~\ref{tab:nnls} also provides the
function values of the individual solvers. It can be seen, while the
SCS solver is considerably faster than the ECOS solver, the solution
computed by the SCS solver is not optimal in experiment (i). The ECOS
solver provides a solution with the same accuracy as GENO but at a
running time that is a few orders of magnitude larger.

\begin{table}[h]
  \centering
  \caption{Running times and function values for the non-negative
    least squares problem.}
  \label{tab:nnls}
  \begin{tabular}{rrlrrrrr}
    \toprule
    m & n & & GENO & ECOS & SCS & Gurobi & Mosek \\
    \midrule
    \multirow{2}{*}{2000} & \multirow{2}{*}{6000} & time & 4.8 & 689.7 & 70.4 & 187.3 & 24.9 \\
    && fval & 0.01306327 & 0.01306327 & 0.07116707 & 0.01306330 & 0.01306343 \\
    \midrule
    \multirow{2}{*}{6000} & \multirow{2}{*}{3000} & time & 0.3 & 3751.3 & 275.5 & 492.9 & 58.4 \\
    && fval & 0.03999098 & 0.03999098 & 0.04000209 & 0.03999100 & 0.03999114 \\
    \bottomrule
  \end{tabular}
\end{table}

\subsection{Symmetric Non-negative Matrix Factorization}

Non-negative matrix factorization (NMF) and its many variants are
standard methods for recommender systems~\cite{adomavicius2005} and
topic modeling~\cite{blei2003,hofmann1999}. It is known as symmetric
NMF, when both factor matrices are required to be identical. Symmetric
NMF is used for clustering problems~\cite{kuang2015} and known to be
equivalent to $k$-means kernel clustering~\cite{ding2005}. Given a
target matrix $T\in\setR^{n\times n}$, symmetric NMF is given as the
following optimization problem
\[
\min_{U}\: \norm{T-UU^\top}_{\mbox{\scriptsize Fro}}^2 \quad\st\: U \geq 0,
\]
where $U\in\setR^{n\times k}$ is a positive factor matrix of rank
$k$. Note, the problem cannot be modeled and solved by CVXPY since
it is non-convex. It has been addressed recently in~\cite{ZhuLLL18} by
two new methods. Both methods are symmetric variants of the
alternating non-negative least squares (ANLS)~\cite{kim2008} and the
hierarchical ALS (HALS)~\cite{cichocki2009} algorithms.

\begin{figure*}[h]
  \centering
  \begin{tabular}{lcr}
    \includegraphics[width=0.31\textwidth]{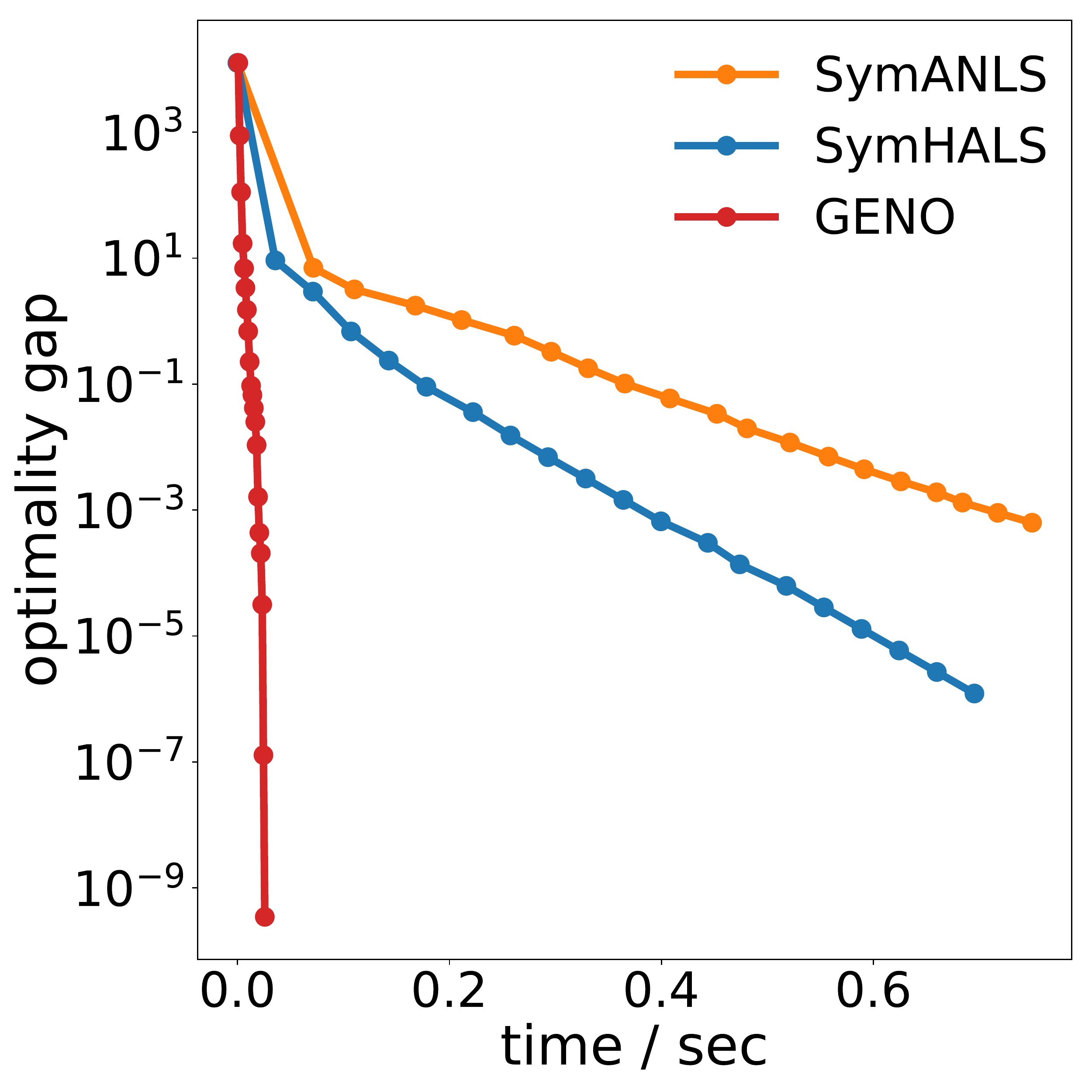} &
    \includegraphics[width=0.31\textwidth]{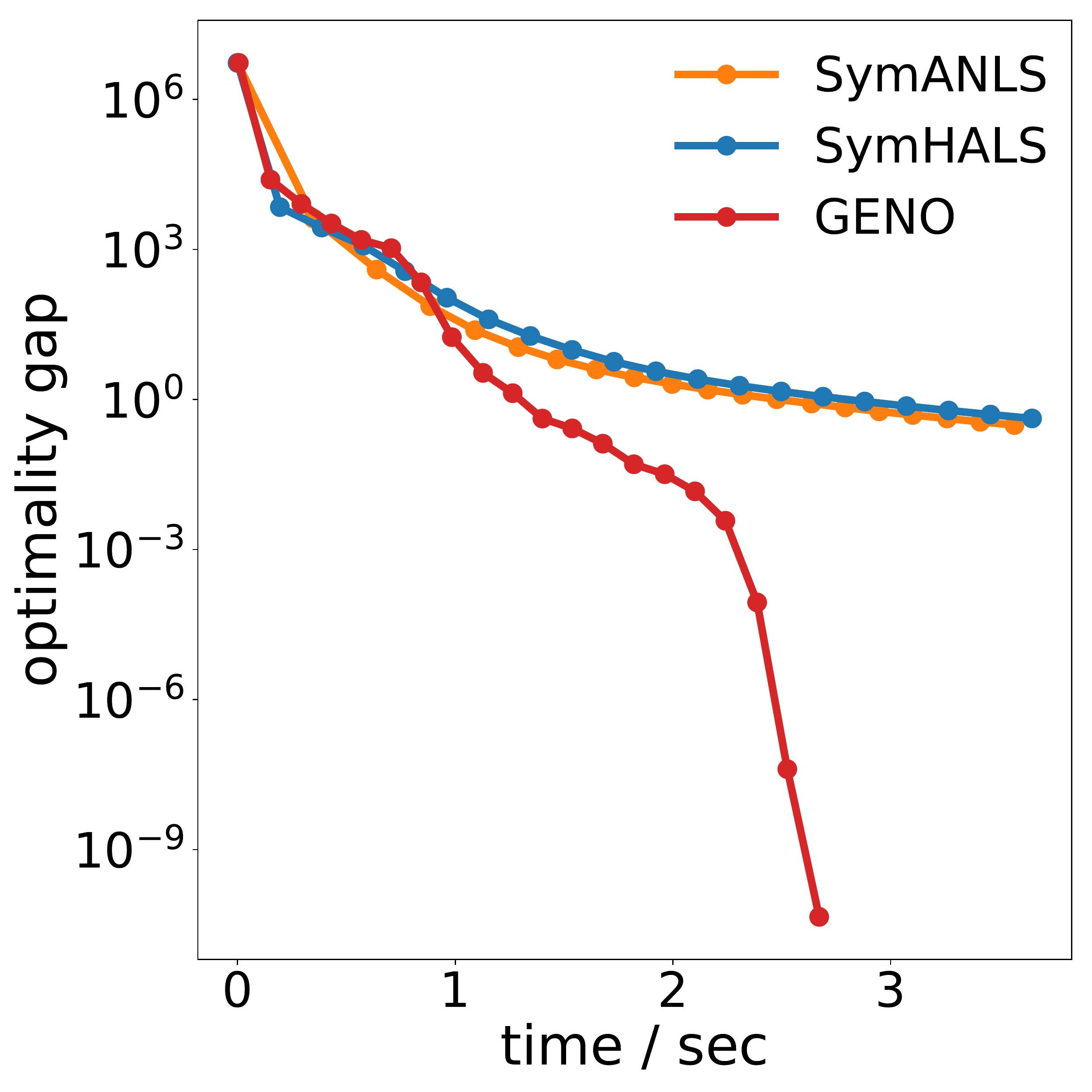}
    \includegraphics[width=0.31\textwidth]{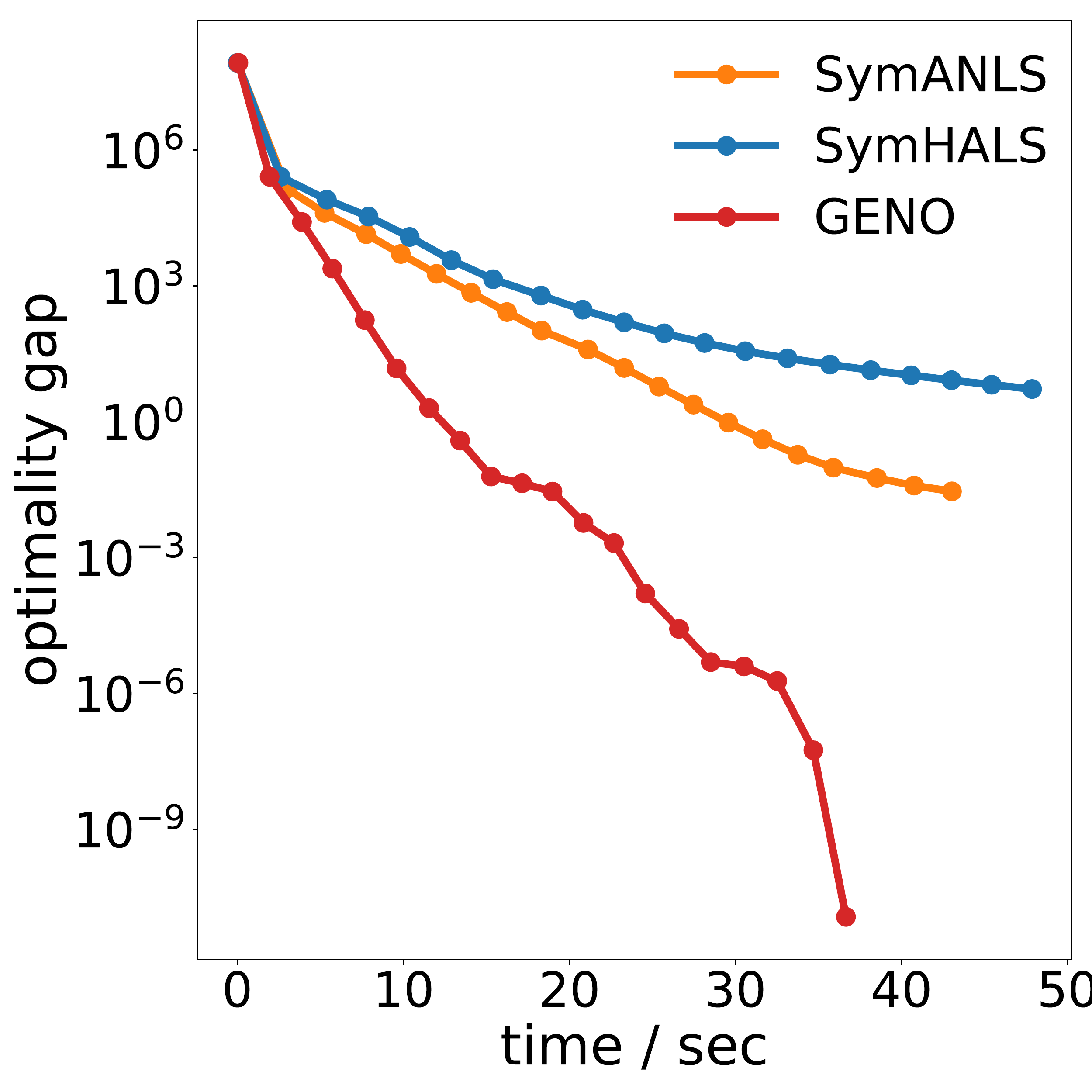}
    \end{tabular}
  \caption{Convergence speed on the symmetric non-negative matrix
    factorization problem for different parameter values. On the left,
    the times for $m = 50, k=5$, in the middle for $m=500, k=10$, and
    on the right for $m=2000, k=15$.}
  \label{fig:SNMF}
\end{figure*}

We compared GENO to both methods. For the comparison we used the code
and same experimental setup as in~\cite{ZhuLLL18}. Random
positive-semidefinite target matrices $X=\hat U\hat U^\top$ of
different sizes were computed from random matrices $\hat
U\in\setR^{n\times k}$ with absolute value Gaussian entries. As can be
seen in Figure~\ref{fig:SNMF}, GENO outperforms both methods (SymANLS
and SymHALS) by a large margin.

\subsection{Non-linear Least Squares}

\begin{figure*}[h]
  \centering
  \begin{tabular}{ccc}
    \includegraphics[width=0.4\textwidth]{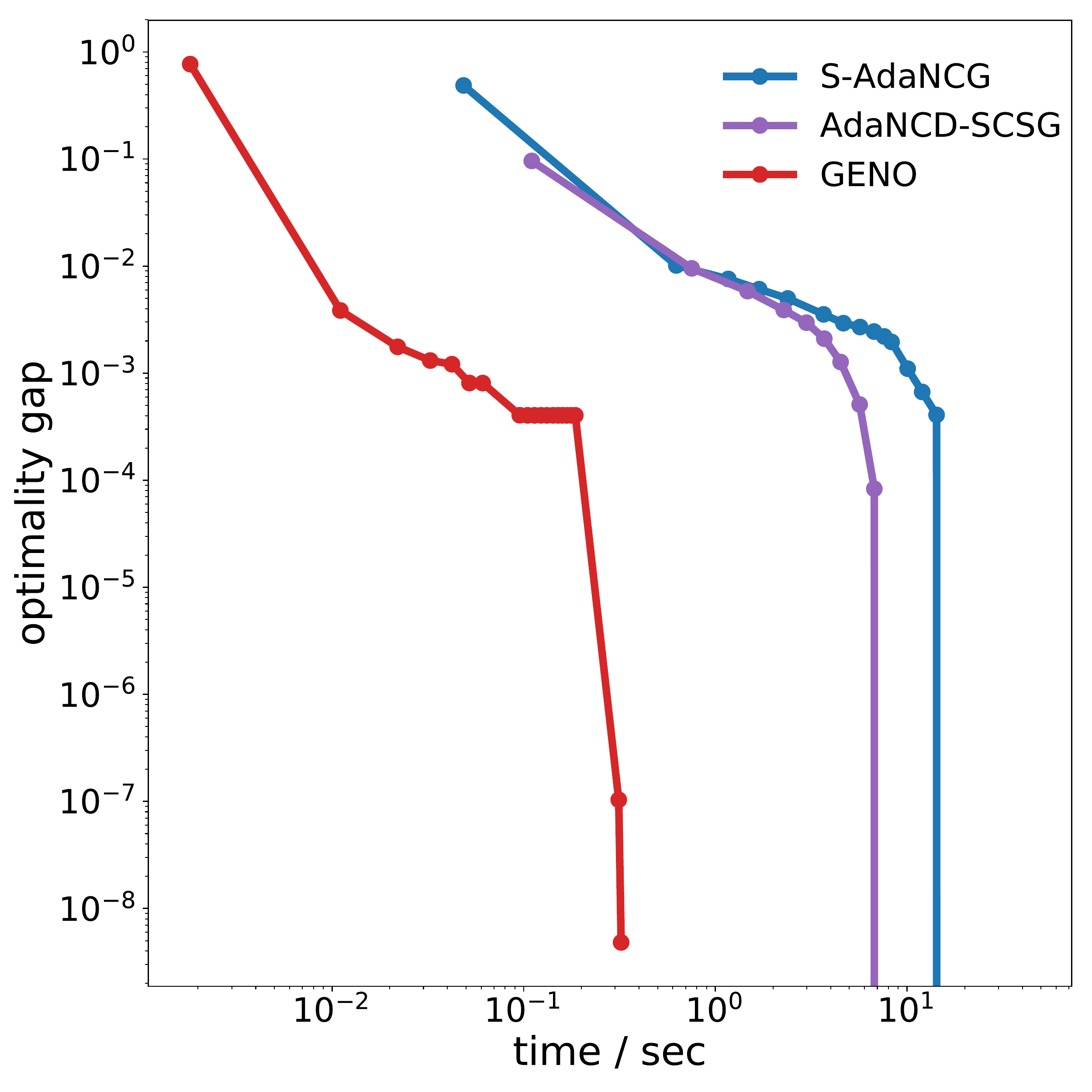} &
    \quad\quad & 
    \includegraphics[width=0.4\textwidth]{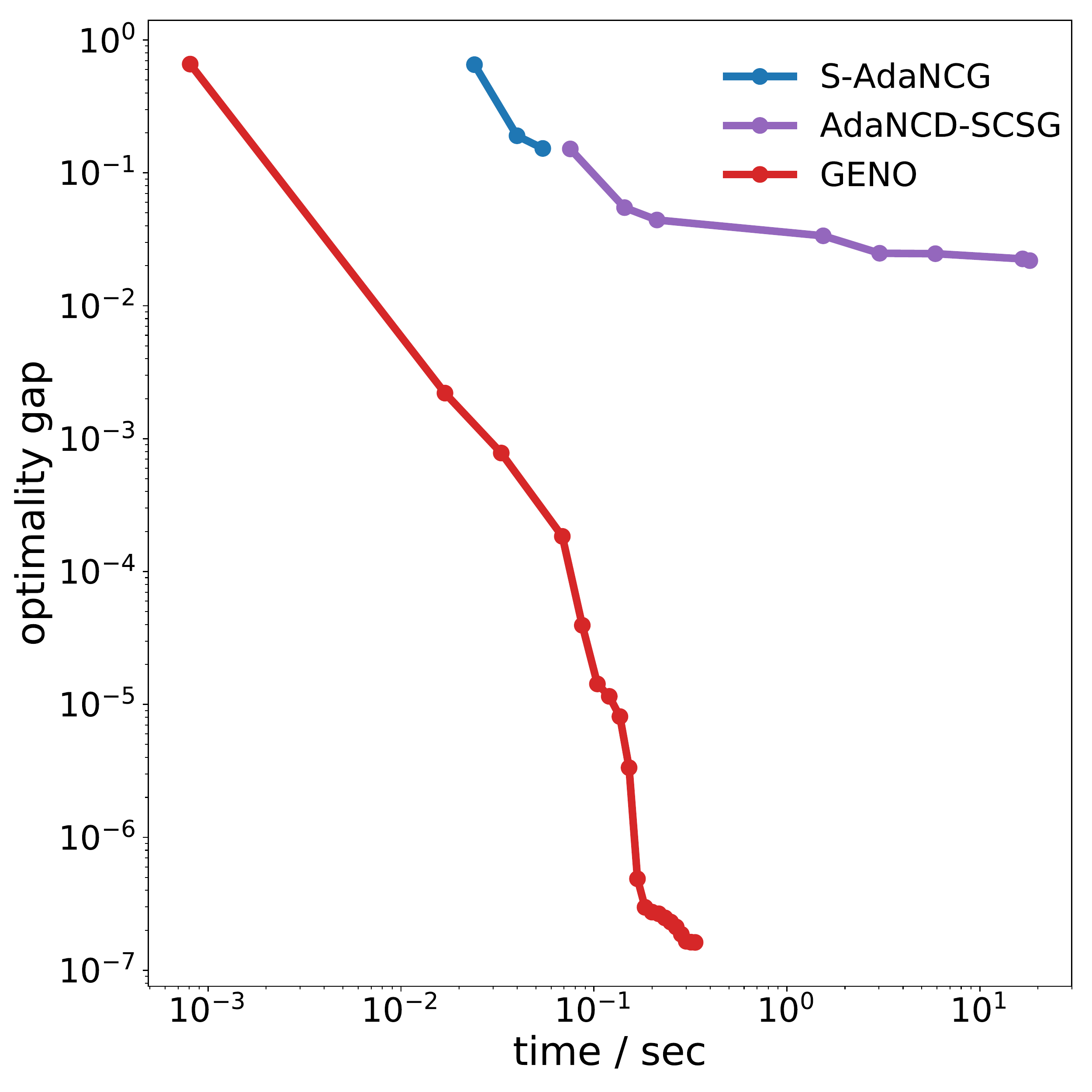}
  \end{tabular}
  \caption{Running times for the non-linear least squares problem. The
    figure on the left shows the running times for the data set
    \texttt{w1a} and on the right for the data set \texttt{a1a}.}
  \label{fig:nonlinearLS}
\end{figure*}

GENO makes use of a quasi-Newton solver which approximates the Hessian
by the weighted sum of the identity matrix and a positive
semidefinite, low-rank matrix. One could assume that this does not
work well in case that the true Hessian is indefinite, i.e., in the
non-convex case. Hence, we also conducted some experiments on
non-convex problems. We followed the same setup and ran the same
experiments as in~\cite{LiuLWYY18} and compared to state-of-the-art
solvers that were specifically designed to cope with non-convex
problems. Especially, we considered the non-linear least squares
problem, i.e., we seek to minimize the function $l(x) =
\norm{\sigma(Ax) - b}_2^2$, where $A\in\setR^{m\times n}$ is a data
matrix, $y\in\{0, 1\}^m$ is a binary label vector, and $\sigma(s) =
1/(1+\exp(-s))$ is the sigmoid function. Figure~\ref{fig:nonlinearLS}
shows the convergence speed for the data set \texttt{w1a} and
\texttt{a1a}. The state-of-the-art specialized solvers that were
introduced in~\cite{LiuLWYY18} are \mbox{S-AdaNCG}, which is a
stochastic adaptive negative curvature and gradient algorithm, and
\mbox{AdaNCD-SCSG}, an adaptive negative curvature descent algorithm
that uses SCSG~\cite{lei2017} as a subroutine. The experiments show
that GENO outperforms both algorithms by a large margin. In fact, on
the data set \texttt{a1a}, both algorithms would not converge to the
optimal solution with standard parameter settings. Again, this problem
cannot be modeled and solved by CVXPY.

\subsection{Compressed Sensing}

In compressed sensing, one tries to recover a sparse signal from a
number of measurements~\cite{CandesT06,donoho2006}. See the recent
survey~\cite{Rani18} for an overview on this topic. The problem can be
reduced to finding the solution to an underdetermined system of linear
equations with minimal $\ell_1$-norm. Hence, it can be written as the
following optimization problem
\begin{equation} \label{eq:cs}
  \begin{array}{rl}
    \min_{x} & \norm{x}_1\\
    \st & Ax= b,
  \end{array}
\end{equation}
where $A\in\setR^{m\times n}$ is a measurement matrix and
$b\in\setR^m$ is the vector of $m$ measurements. Note, that this
problem is a constrained problem with a non-differentiable objective
function. It is known that when matrix $A$ has the restricted isometry
property and the true signal $x^*$ is sparse, then
Problem~\eqref{eq:cs} recovers the true signal with high probability,
if the dimensions $m$ and $n$ are chosen
properly~\cite{candes2005}. There has been made considerable progress
in designing algorithms that come with convergence
guarantees~\cite{chin2013,christiano2011}. Very recently,
in~\cite{Ene19} a new and efficient algorithm based on the iterative
reweighted least squares (IRLS) technique has been proposed. Compared
to previous approaches, their algorithm is simple and achieves the
state-of-the-art convergence guarantees for this problem.

\begin{figure*}[t]
  \centering
  \begin{tabular}{ccc}
    \includegraphics[width=0.4\textwidth]{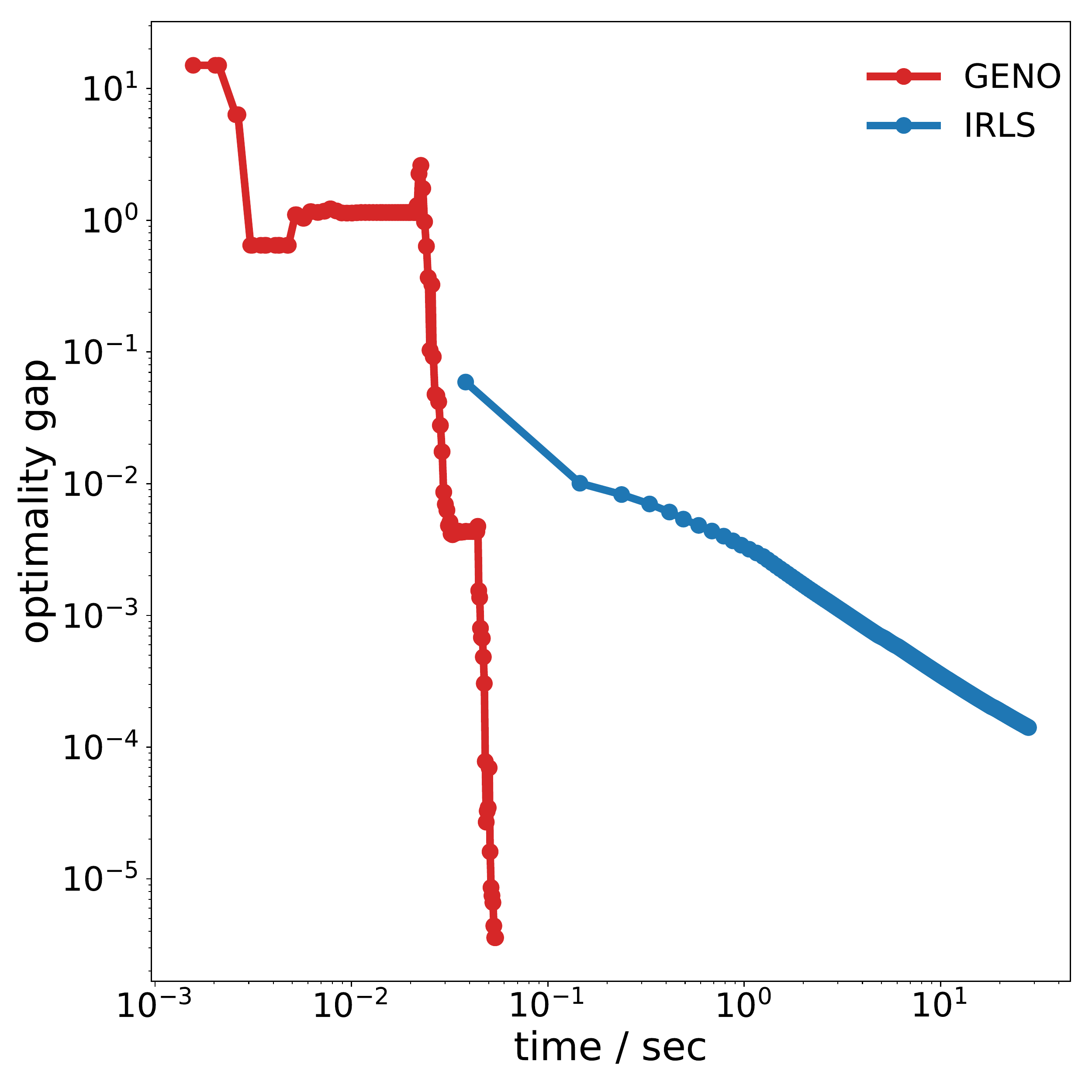} &
    \quad\quad &
    \includegraphics[width=0.4\textwidth]{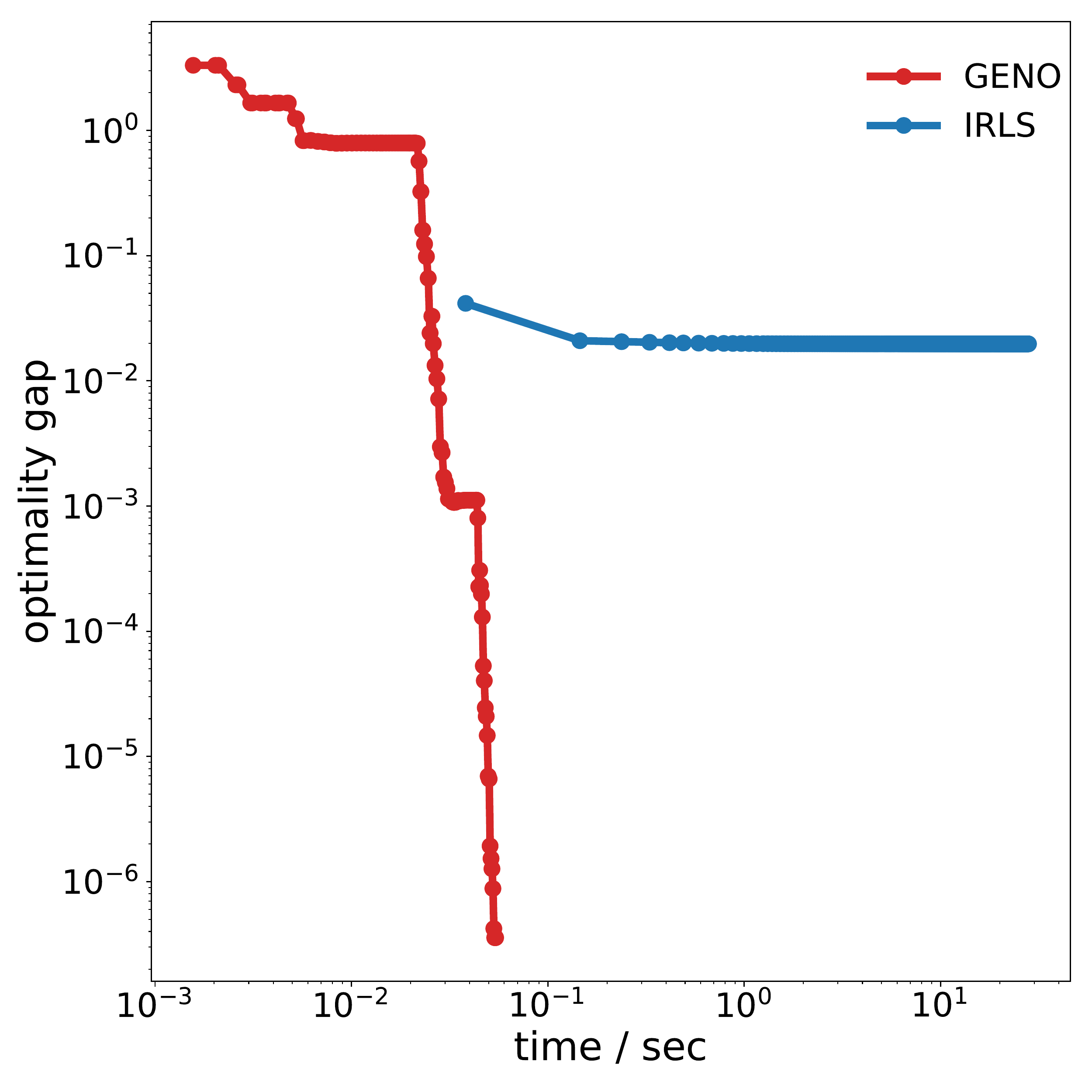}
  \end{tabular}
  \caption{Running times for the compressed sensing problem. The
    figure on the left shows the convergence to the optimal objective
    function value and the figure on the right shows the norm of the
    constraint violation of the iterate $x^{(k)}$, i.e.,
    $\norm{Ax^{(k)}-b}_2$.}
  \label{fig:cs}
\end{figure*}

We used the same setup and random data set as in~\cite{Ene19} and ran
the same experiment. The measurement matrix $A\in\setR^{150\times
  200}$ had been generated randomly, such that all rows are
orthogonal. Then, a sparse signal $x^*$ with only 15 non-zero entries
had been chosen and the corresponding measurement vector $b$ had been
computed via $b=Ax^*$. We compared to their IRLS algorithm with the
long-steps update scheme. Figure~\ref{fig:cs} shows the convergence
speed speed towards the optimal function value as well as the
convergence towards feasibility. It can be seen that the solver
generated by GENO outperforms the specialized, state-of-the-art IRLS
solver by a few orders of magnitude.

\section{Conclusions}
\label{sec:conclusions}

While other fields of applied research that heavily rely on
optimization, like operations research, optimal control and deep
learning, have adopted optimization frameworks, this is not the case
for classical machine learning. Instead, classical machine learning
methods are still mostly accessed through toolboxes like scikit-learn,
Weka, or MLlib. These toolboxes provide well-engineered solutions for
many of the standard problems, but lack the flexibility to adapt the
underlying models when necessary. We attribute this state of affairs
to a common belief that efficient optimization for classical machine
learning needs to exploit the problem structure. Here, we have
challenged this belief. We have presented GENO the first general
purpose framework for problems from classical machine learning. GENO
combines an
easy-to-read modeling language with a general purpose
solver. Experiments on a variety of problems from classical machine
learning demonstrate that GENO is as efficient as established
well-engineered solvers and often outperforms recently published
state-of-the-art solvers by a good margin. It is as flexible as
state-of-the-art modeling language and solver frameworks, but
outperforms them by a few orders of magnitude.

\section*{Acknowledgments}
S\"oren Laue has been funded by Deutsche Forschungsgemeinschaft (DFG) under grant LA~2971/1-1.

\bibliographystyle{plain}
\bibliography{geno}

\clearpage
\newpage
\appendix

\section*{Appendix}
\section{GENO Models for all Experiments}
\newcommand*{\languagesize}{\footnotesize}
\newcommand*{\languagepagewidth}{0.48\textwidth}
\begin{figure}[!htb]
\begin{minipage}{\languagepagewidth}
\begin{Verbatim}[frame=single, fontsize=\languagesize]
parameters
   Matrix X
   Scalar c
   Vector y
variables
   Scalar b
   Vector w
min
   norm1(w) + c
   * sum(log(exp((-y) .* (X*w
   + vector(b))) + vector(1)))     
\end{Verbatim}
\vskip -2ex \caption{$\ell_1$-regularized Logistic Regression}
\label{fig:modeling:LR1}
\vspace{0.3cm}
\end{minipage}
\begin{minipage}{\languagepagewidth}
\begin{Verbatim}[frame=single, fontsize=\languagesize]
parameters
   Matrix X
   Scalar c
   Vector y
variables
   Vector w
min
   0.5 * w' * w
   + c * sum(log(exp((-y) .* (X * w))
   + vector(1)))

\end{Verbatim}
\label{fig:modeling:LR2}
\vskip -2ex\caption{$\ell_2$-regularized Logistic Regression}
\vspace{0.3cm}
\end{minipage}
\begin{minipage}{\languagepagewidth}
\begin{Verbatim}[frame=single, fontsize=\languagesize]
parameters
   Matrix K symmetric
   Scalar c
   Vector y
variables
   Vector a
min
   0.5 * (a.*y)' * K * (a.*y) - sum(a)
st
   a >= 0
   y' * a == 0
\end{Verbatim}
\label{fig:modeling:SVM}
\vskip -2ex\caption{Support Vector Machine}
\end{minipage}
\vspace{0.3cm}
\begin{minipage}{\languagepagewidth}
\begin{Verbatim}[frame=single, fontsize=\languagesize]
parameters
   Matrix X
   Scalar a1
   Scalar a2
   Scalar n
   Vector y
variables
   Vector w
min
   n * norm2(X*w - y).^2
   + a1 * norm1(w) + a2 * w' * w
\end{Verbatim}
\vskip -2ex\caption{Elastic Net}
\label{fig:modeling:ElasticNet}
\end{minipage}
\vspace{0.3cm}
\begin{minipage}{\languagepagewidth}
\begin{Verbatim}[frame=single, fontsize=\languagesize]
parameters
   Matrix A
   Vector b
variables
   Vector x
min
   norm2(A*x - b).^2
st
   x > 0
\end{Verbatim}
\label{fig:modeling:NNLS}
\vskip -2ex\caption{Non-negative Least Squares}
\end{minipage}
\vspace{0.5cm}
\begin{minipage}{\languagepagewidth}
\begin{Verbatim}[frame=single, fontsize=\languagesize]
parameters
   Matrix X symmetric
variables
   Matrix U
min
   norm2(X - U*U').^2
st
   U >= 0
  
\end{Verbatim}
\label{fig:modeling:SNNMF}
\vskip -2ex\caption{Symmetric NMF}
\end{minipage}
\vspace{-0.4cm}
\begin{minipage}{\languagepagewidth}
\begin{Verbatim}[frame=single, fontsize=\languagesize]
parameters
   Matrix X
   Scalar s
   Vector y
variables
   Vector w
min
   s * norm2(y - 0.5
   * tanh(0.5 * X * w)
   + vector(0.5)).^2
\end{Verbatim}
\label{fig:modeling:NLLS}
\vskip -2ex\caption{Non-linear Least Squares}
\end{minipage}
\hspace{0.4cm}
\begin{minipage}{\languagepagewidth}
\begin{Verbatim}[frame=single, fontsize=\languagesize]
parameters
   Matrix A
   Vector b
variables
   Vector x
min
   norm1(x)
st
   A*x == b

\end{Verbatim}
\label{fig:modeling:CS}
\vskip -2ex\caption{Compressed Sensing}
\end{minipage}
\end{figure}

\newpage
\section{Summary of Solvers}
\begin{table}[h]
	\centering
	\caption{Summary of all solvers that were used in the experiments.}
	\label{tab:solvers}
	\begin{tabular}{lll}
		\toprule
		Name & Type & Reference \\
		\midrule
		\addlinespace[0.5em]
		GENO & quasi-Newton w/ augmented Lagrangian & this paper\\
		\addlinespace[0.5em]
		OSQP & ADMM & \cite{osqp} \\
		SCS & ADMM & \cite{scs}\\
		ECOS & interior point & \cite{Ecos1}\\
		Gurobi & interior point & \cite{gurobi}\\
		Mosek & interior point & \cite{mosek}\\
		LIBSVM & Frank-Wolfe (SMO) & \cite{ChangL01}\\
		\multirow{2}{*}{LIBLINEAR} &  conjugate gradient  +  & \multirow{2}{*}{\cite{FanCHWL08,ZhuangJYL18}}\\ & dual coordinate descent & \\
		SAGA & SGD & \cite{DefazioBL14} \\
		SDCA & SGD & \cite{Shalev-Shwartz13}\\
		catalyst SDCA & SGD & \cite{LinMH15}\\
		Point-SAGA & SGD & \cite{Defazio16}\\
		glmnet & dual coordinate descent & \cite{FriedmanHT09}\\
		Lawson-Hanson & direct linear equations solver w/ active set  & \cite{lawson95}\\
		projected gradient descent & proximal algorithm & \cite{Slawski13} \\
		primal-dual interior point  w/ & \multirow{2}{*} {interior point} & \multirow{2}{*}{\cite{Slawski13}}\\
		preconditioned conjugate gradient \\
		subspace Barlizai-Borwein & quasi-Newton & \cite{kim13}\\
		Nesterov's method & accelerated gradient descent & \cite{Nesterov83}\\
		SymANLS & block coordinate descent & \cite{ZhuLLL18}\\
		SymHALS & block coordinate descent & \cite{ZhuLLL18}\\
		S-AdaNCG & SGD& \cite{LiuLWYY18}\\
		AdaNCD-SCSG & SGD & \cite{LiuLWYY18}\\
		IRLS & IRLS w/ conjugate gradient method & \cite{Ene19} \\
		\bottomrule
	\end{tabular}
\end{table}

\end{document}